%% file: main.tex
\documentclass[10pt,twocolumn,letterpaper]{article}

\usepackage[pagenumbers]{cvpr} 

\input{preamble}
\definecolor{cvprblue}{rgb}{0.21,0.49,0.74}
\usepackage[pagebackref,breaklinks,colorlinks,allcolors=cvprblue]{hyperref}
\usepackage{array}
\def\paperID{ } 
\def\confName{CVPR}
\def\confYear{2026}

\title{Perception Characteristics Distance: Measuring Stability and Robustness of Perception System in Dynamic Conditions under a Certain Decision Rule}

\author{
Boyu Jiang$^{1}$\thanks{Boyu Jiang and Liang Shi contributed equally to this work.}\hspace{0.7em}
Liang Shi$^{1,2}$\footnotemark[1]\hspace{0.7em}
Zhengzhi Lin$^{1}$\hspace{0.7em}
Lanxin Xiang$^{1}$\hspace{0.7em}
Loren Stowe$^{2}$\hspace{0.7em}
Feng Guo$^{1,2}$\thanks{Corresponding author.}\\[0.4em]
$^{1}$Department of Statistics, Virginia Tech\hspace{1.5em}
$^{2}$Virginia Tech Transportation Institute\\[0.2em]
{\tt\small feng.guo@vt.edu}
}

\begin{document}
\maketitle
\input{sec/0_abstract}    
\input{sec/1_intro}
\input{sec/2_related_works}
\input{sec/3_dataset}

\input{sec/4_PCD}

\input{sec/5_experiment}
\input{sec/6_conclusion}
{
    \small
    \bibliographystyle{ieeenat_fullname}
    \bibliography{main}
}


\input{sec/X_suppl}

\end{document}

%% file: sec/0_abstract.tex
\begin{abstract}

The safety of autonomous driving systems (ADS) depends on accurate perception across distance and driving conditions. The outputs of AI perception algorithm are stochastic, which have a major impact on decision making and safety outcomes, including time-to-collision estimation. However, current perception evaluation metrics do not reflect the stochastic nature of perception algorithms. We introduce the \textbf{Perception Characteristics Distance (PCD)}, a novel metric incorporating model output uncertainty as represented by the farthest distance at which an object can be \textbf{reliably} detected. To represent a system’s overall perception capability in terms of reliable detection distance, we used the averaging PCD values across multiple detection quality and probabilistic thresholds produces the \textbf{average PCD (aPCD)}. For empirical validation, we present the \textbf{SensorRainFall} dataset, collected on the Virginia Smart Road using a sensor-equipped vehicle (cameras, radar, and LiDAR) controlled under different weather (clear and rainy) and illumination conditions (daylight, streetlight, and nighttime). The dataset includes ground-truth distances, bounding boxes, and segmentation masks for target objects. Experiments with state-of-the-art models show that aPCD captures meaningful differences across weather, daylight, and illumination conditions, which traditional evaluation metrics fail to reflect. PCD provides an uncertainty-aware measure of perception performance, supporting safer and more robust ADS operation, while the SensorRainFall dataset offers a valuable benchmark for evaluation. The \textit{SensorRainFall} dataset is publicly available at \url{https://www.kaggle.com/datasets/datadrivenwheels/sensorrainfall}, and the evaluation code is available at \url{https://github.com/datadrivenwheels/PCD_Python}
\end{abstract}

%% file: sec/1_intro.tex
\section{Introduction}
\label{sec:intro}

Perception systems, particularly object detection and segmentation, play a central role in intelligent decision-making for AI-powered vision applications. In autonomous driving systems (ADS) and advanced driver-assistance systems (ADAS), they enable real-time understanding and interaction with complex, dynamic environments. However, standard evaluation metrics such as average precision (AP), Intersection-over-Union (IoU), and the F1 score are rely on static, frame-level assessments. These metrics overlook the temporal and spatial continuity inherent to real-world driving. These metrics provide limited insight into the robustness of perception systems under dynamic operating conditions.

\begin{figure}
  \centering
  \includegraphics[scale=0.28]{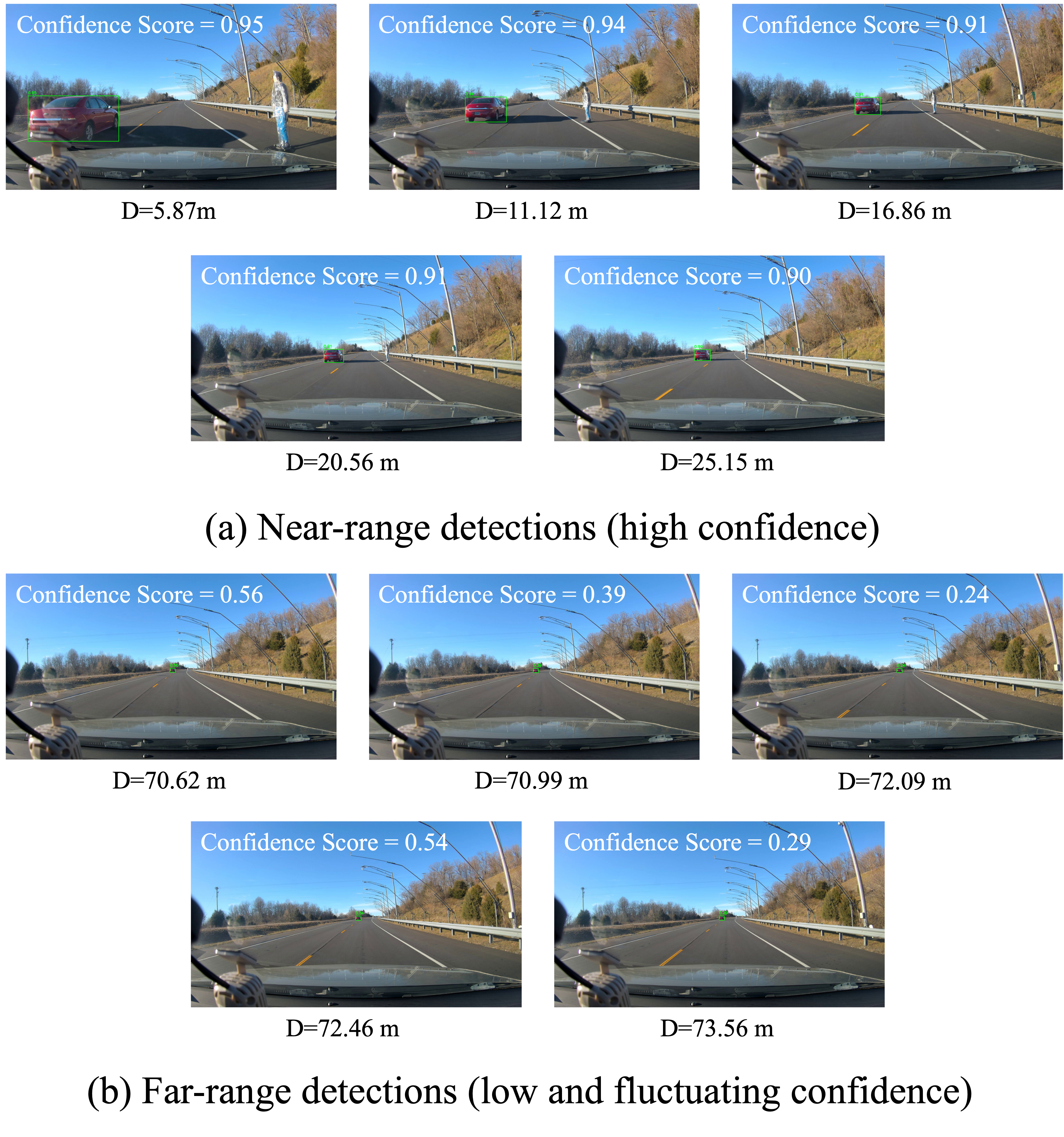}
  \caption{Comparison of detection confidence at near vs. far ranges. Near-range object detections with consistently high confidence scores ($\geq$ 0.90). Far-range detections exhibit unstable and fluctuating confidence, including brief drops as low as 0.24. Despite visual continuity of the object, detection reliability degrades with increasing distance.}
  \label{fig:daylight_example}
\end{figure}

In practical ADAS and ADS deployments, vehicle control logic often relies on threshold-based detection, where an object is considered detected if the model’s confidence score exceeds a fixed threshold. This binary approach does not fully capture the stochastic and distance-dependent variability of perception outputs. Figure \ref{fig:daylight_example} illustrates this limitation using the performance of a YOLOX-based object detector at near range (a) and far range (b). As the ego vehicle follows a leading car on the highway, detection confidence fluctuates considerably at farther distances ($\geq$ 70 m, Figure \ref{fig:daylight_example} (b)). These fluctuations challenge downstream control modules that depend on stable, high-confidence detections for safe acceleration or braking. Applying a fixed threshold (e.g., confidence score $\geq$ 0.4) leads to inconsistent detection decisions, making the maximum reliable detection range uncertain. This motivates the development of a dynamic evaluation framework that captures spatial variability and distance-dependent reliability of perception systems under realistic conditions.

To address these challenges, we propose a novel evaluation metric—Perception Characteristics Distance (PCD)—that quantifies the maximum distance within which a perception system can consistently produce reliable detections at a given detection threshold. Unlike traditional static metrics, PCD accounts for the spatial dynamics of perception by statistically estimating the mean and variance of detection confidence as a function of distance. This probabilistic formulation enables PCD to robustly capture performance variability over time and space, offering a more accurate and context-aware assessment of system reliability under real-world conditions. To enable a comprehensive comparison, we define the average PCD (aPCD) as the average of PCD values computed over a range of detection thresholds.

For benchmarking, we present the SensorRainFall dataset, which, to the best of the authors’ knowledge, is the only publicly available dataset collected in a highly controlled environment specifically designed for perception evaluation in ADS and ADAS. Data were collected on the Virginia Smart Roads, a facility capable of simulating diverse weather conditions. All other environmental factors were held constant, while only the weather (clear and rainy) and illumination conditions (daylight, nighttime, and nighttime with streetlights) were varied across scenarios. The dataset includes both object-level and pixel-level annotations across multiple sensor modalities (camera, radar, and LiDAR). The dataset includes both object-level and pixel-level annotations across multiple sensor modalities (camera, radar, LiDAR). Each image was manually reviewed and annotated with ground-truth bounding boxes, instance segmentation masks, and precise distance measurements for two target objects: a red sedan and a dummy pedestrian.



%% file: sec/2_related_works.tex
\section{Related works}
\label{litrew}

\noindent \textbf{Video Datasets Supporting Autonomous Driving Perception.} Video datasets are vital for advancing autonomous driving perception by capturing real-world variability. nuScenes offers rich multimodal data for complex urban environments \citep{caesar2020nuscenes}, while KITTI remains a foundational benchmark for object detection and tracking \citep{Geiger2013IJRR}. SHRP 2 provides high-frequency video and telemetry from naturalistic driving, enabling analysis of driver behavior and risk factors \citep{hankey2016}. BDD100K delivers large-scale annotated driving videos across diverse weather, lighting, and locations \citep{yu2020bdd100k}. Brain4Cars focuses on driver maneuver anticipation using both in-cabin and exterior views \citep{jain2016brain4cars}. However, these naturalistic datasets lack controlled environments, which may limit experimental consistency in certain research settings.

\noindent \textbf{Perception Evaluation Metrics.} In perception evaluation, metrics such as precision, recall, F1 score, mAP, and mIoU remain widely used, though they treat each detection independently. To address these limitations, Can and Cigla \citep{oksuz2018localization} introduced the Localization Recall Precision (LRP) Error and its optimized variant (oLRP) as an alternative to AP, providing a more discriminative evaluation by jointly accounting for localization accuracy, false positives, and false negatives in object detection. Subsequent extensions have incorporated temporal aspects; for instance, Zhang et al. \citep{zhang2016stability} enhance mAP with stability components for video detection, while Mao et al. \citep{mao2019delay} propose the Average Delay (AD) metric to capture detection latency under false alarm constraints. Beyond temporal behavior, recent work has also emphasized uncertainty modeling: Hall et al. \citep{hall2020probabilistic} introduce the Probabilistic Detection Quality (PDQ) metric, which jointly captures spatial and semantic uncertainty, while Feng et al. \citep{feng2021labels} propose Jaccard IoU with uncertainty (JIoU) to address annotation noise and imperfect ground-truth localization.
In parallel, Rezatofighi et al. \citep{rezatofighi2019generalized} propose the Generalized Intersection over Union (GIoU) metric to address IoU’s failure for non-overlapping bounding boxes. While these advances improve evaluation along spatial, temporal, and uncertainty dimensions, PCD uniquely emphasizes distance-dependent reliability and fluctuations in detection confidence, providing a metric specifically suited for evaluating perception in ADS/ADAS contexts.



%% file: sec/3_dataset.tex
\section{SensorRainfall Dataset}
\label{dataset}

The SensorRainfall dataset was developed under both controlled and dynamic test environments to assess the impact of rainfall intensity on sensor performance. This dataset is particularly valuable for analyzing the influence of precipitation on perception capabilities and overall system-level performance of ADAS and ADS.

Data collection was conducted on the Virginia Smart Roads, a facility capable of simulating diverse weather conditions, including rainfall, over a test track that is more than 200 meters in length \citep{gibson2015virginia}. The testing scenario emulated Forward Collision Warning (FCW) and Automatic Emergency Braking (AEB) systems.  Four environmental conditions were recorded: Clear daylight, Rainy daylight, Rainy night, and Rainy night under streetlights, with all rainy conditions conducted under a controlled rainfall rate of 64\,mm/h \citep{cowan2024investigation}.

\begin{figure}
  \centering
  \includegraphics[scale=0.29]{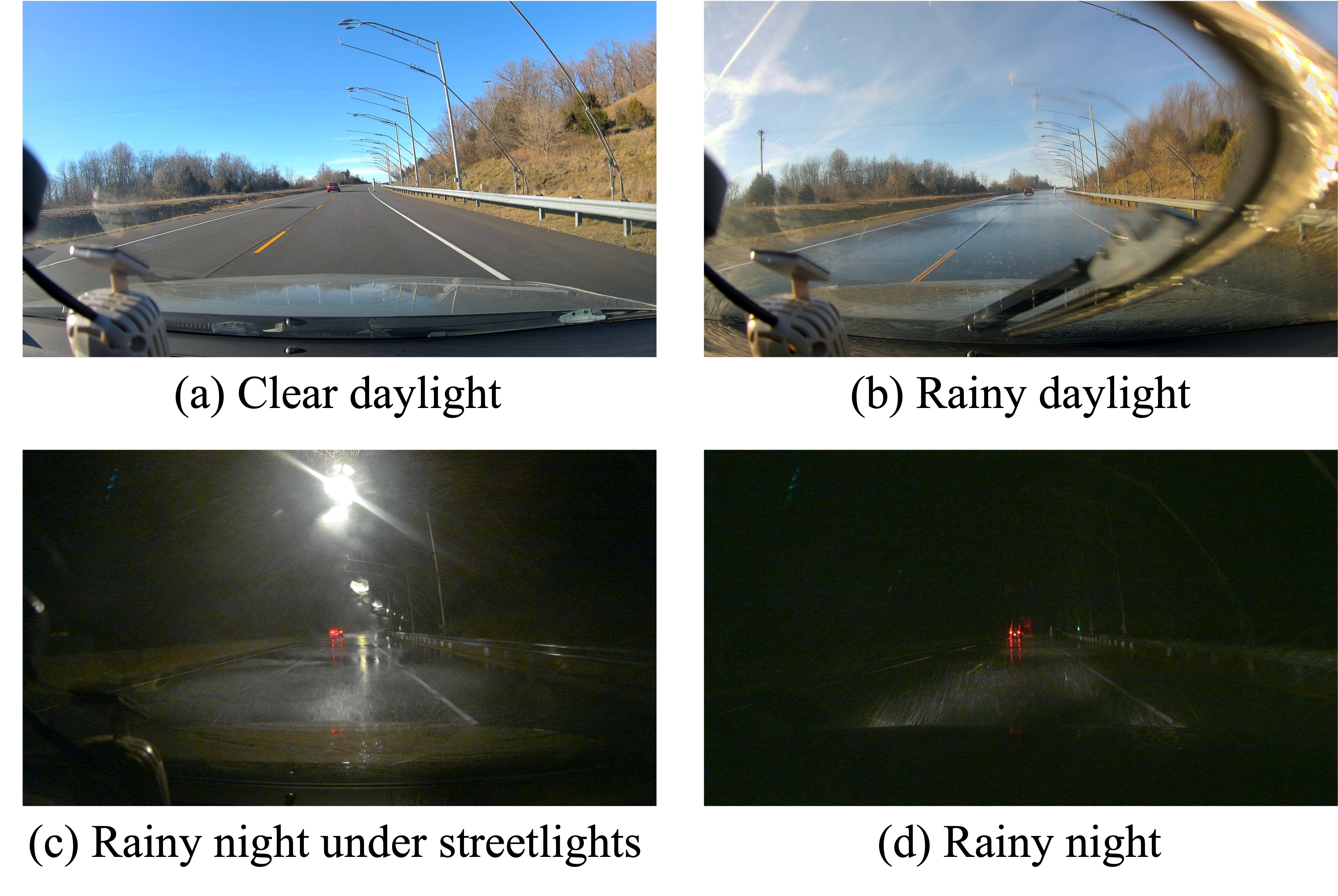}
  \caption{SensorRainfall dataset example.}
  \label{fig:example data}
\end{figure}

The dataset contains 1,231 sequential front-view images captured from the ego vehicle, featuring two target objects: a red sedan and a dummy pedestrian. These images were recorded at distances ranging from 4m to 250m under the four environmental conditions. All images were collected at a resolution of 1920$\times$1080. Representative sample images are shown in Figure~\ref{fig:example data}. Each image was manually reviewed by trained researchers and annotated with ground-truth bounding boxes and pixel-level segmentation masks for the red sedan and dummy pedestrian.



%% file: sec/4_PCD.tex
\section{Perception Characteristics Distance}
\label{method}

\begin{figure}
  \centering
  \includegraphics[scale=0.33]{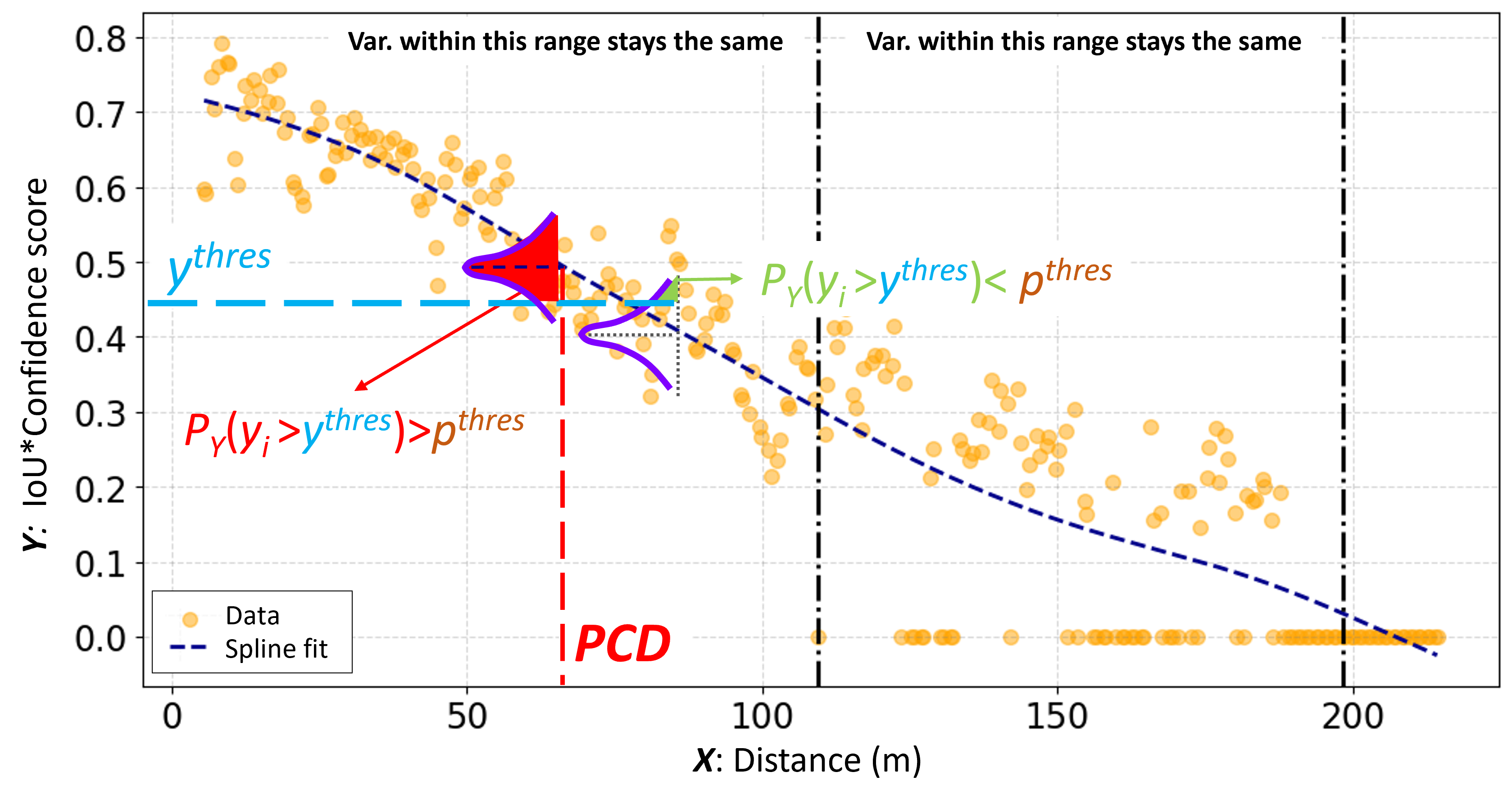}
  \caption{Conceptual illustration of PCD. 
  }
  \label{fig:pipeline}
\end{figure}

The PCD is defined as the maximum distance $x_i$ for which the probability $P_Y(y_i > y^{thres})$ is greater than a predefined probabilistic threshold $p^{thres}$, as illustrated in Figure~\ref{fig:pipeline}.  The horizontal axis ($X$) represents the distance between the ego vehicle and the target object. The vertical axis ($Y$) denotes the IoU$\times$confidence score, a joint measure of spatial accuracy and prediction certainty (see Section \ref{why Iou*Confidence}). As the target object moves farther from the ego vehicle, the perception algorithm’s performance declines, reflected by decreasing IoU$\times$confidence values. Additionally, noticeable variation exists among images captured at similar distances, revealing the inherent stochasticity of perception outputs. PCD captures both of these critical behaviors: (1) the distance-dependent degradation of perception performance and (2) the uncertainty in decision-making caused by fluctuating model outputs.

\subsection{Computation of PCD}

The PCD computation involves two main steps. First, significant variance change points along the $X$-axis are detected to determine the segment boundaries. Second, the PCD is identified by integrating detection confidence, spatial accuracy, and the uncertainty captured within these segments.

As illustrated in Figure \ref{fig:pipeline}, the dynamic determination of PCD begins by analyzing the IoU~$\times$~confidence score across distance and identifying $M$ significant variance change points, denoted as $x_{\tau_1}, x_{\tau_2}, \ldots, x_{\tau_M}$. These change points divide the data into segments exhibiting distinct statistical properties. Within each segment (i.e., between two consecutive variance change points), the IoU~$\times$~confidence score at distance $ x_i$, denoted as $ y_i$,  is assumed to follow a normal distribution:
\begin{align}
    y_i \sim \mathcal{N}&(\mu_i, \sigma_{i}^2); \mu_i = f(x_i), \sigma_{i}=\text{std}(y_A)
\end{align}


where $\mu_i$ is modeled by a spline regression function $f(\cdot)$, $\sigma_i$ is estimated by the standard deviation (std) of $y$s in segment $A$ = $\{j | \tau_m \le j  \le \tau_{m+1}, m=1,2,...,M-1 \}$. As shown in Figure ~\ref{fig:pipeline}, there are two change points, which separates the distance range into three segments.

The PCD is defined as the maximum distance $x_i$ where the probability that $y_i$ exceeds a detection quality threshold $y^{thres}$ is above a probabilistic threshold  $p^{thres}$:
\begin{align}
    PCD &= max \{x_i| P_Y(y_i > y^{thres}) > p^{thres}\}
\end{align}
In this formulation, PCD incorporates both the expected detection quality $\mu_i$, and the associated uncertainty, represented by $\sigma_i$. By adapting to local statistical variations through identified variance change points, PCD establishes a principled, probabilistic cutoff distance beyond which the perception system can no longer be considered reliable.

In practical usage, users can adjust the thresholds $y^{thres}$ and $p^{thres}$ based on their specific detection requirements. For applications requiring higher perception quality, a larger $y^{thres}$ or higher $p^{thres}$ may be chosen. Conversely, in more tolerant scenarios, users may lower these values to extend the PCD while accepting reduced precision. For example, for the scenario of emergency brake for vulnerable road users, a lower threshold is desired as the consequence of missing an object is severe.    

To comprehensively evaluate the perception system across varying thresholds, we compute the volume under the PCD surface over combinations of $p^{thres}$ and $y^{thres}$. This aggregated metric is referred to as the average PCD (aPCD). Similar to how AUC summarizes classifier performance by calculating the area under the ROC curve, which plots the true positive rate against the false positive rate across different thresholds, aPCD captures the overall perception characteristics of a perception system across ranges of two thresholds.

\begin{figure}[htpb]
    \centering
    \subfloat[ ]{%
        \label{fig:base-cam1_Grounding_DINO_work_dist}
        \includegraphics[width=0.42\textwidth]{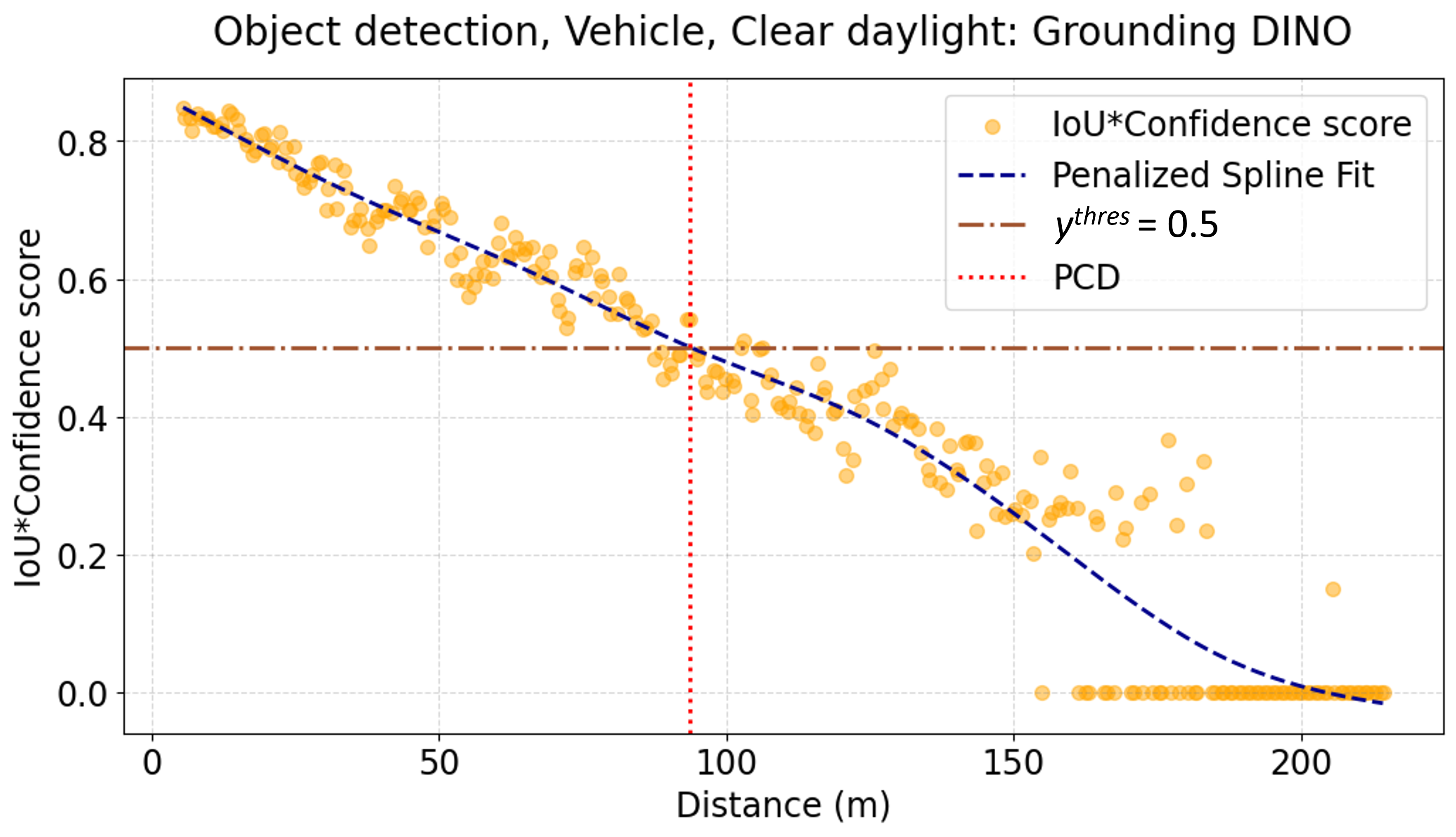}
    }\\[1em]
    \subfloat[ ]{%
        \label{fig:rain-cam1_Grounding_DINO_work_dist}
        \includegraphics[width=0.42\textwidth]{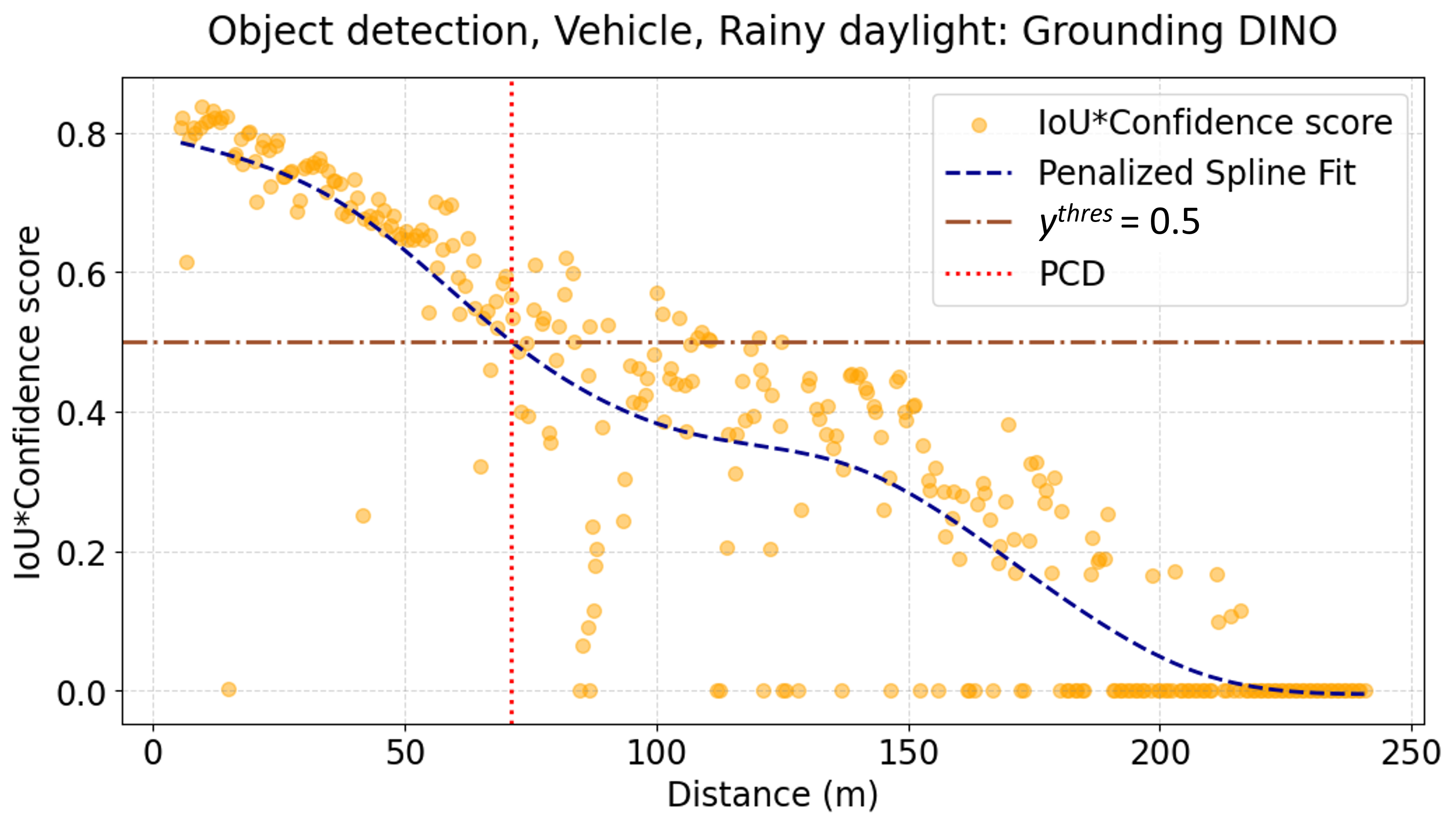}
    }
    \caption{PCD of Grounding DINO for vehicle detection under clear and rainy daylight (threshold: $p^{thres}$ = 0.5 and $y^{thres}$ = 0.5).}
    \label{fig:work_distance} 
\end{figure}

Figure~\ref{fig:work_distance} illustrates the PCD of Grounding~DINO under clear and rainy daylight conditions. As distance increases, the IoU$\times$confidence scores decline. The vertical red dashed lines mark the corresponding PCD values for each condition. Under rainy daylight, detection performance deteriorates sooner, yielding a substantially shorter PCD than in clear daylight.

\subsection{\texorpdfstring{Why IoU $\times$ Confidence score}{IoU x Confidence score}}
\label{why Iou*Confidence}

PCD is derived from the joint consideration of two key aspects of perception quality: spatial accuracy and prediction confidence. Confidence scores indicate how certain a model is about the presence of an object. IoU captures localization accuracy but ignores the model’s confidence in its own prediction. By combining these two measures, IoU $\times$ confidence score (1) captures both the quality and certainty of each detection, (2) provides a more reliable assessment of perception stability across distances, and (3) enables a fair evaluation of detection reliability under varying and dynamic conditions.

\subsection{Detection of variance change points}

To estimate the variance of IoU $\times$ confidence score at each distance point, it is necessary to identify segments where the variance remains approximately constant. This requires detecting variance change points—specific locations in the sequential data where the variance shifts significantly \citep{hawkins2005}. These change points partition the data into segments with distinct levels of variance. Detecting them typically involves first modeling the mean structure of the data using regression techniques \citep{niu2016}, allowing the underlying variance pattern to be isolated and analyzed more accurately.

Splines are particularly effective for this initial regression fitting due to their ability to adaptively capture non-linear patterns without imposing rigid parametric assumptions \citep{perperoglou2019review}. Unlike fixed-form linear models, spline regression decomposes the predictor space into locally smooth polynomial segments joined at optimized knot locations \citep{acharjee2022polynomial}, ensuring mean estimation in complex temporal or spatial relationships. This flexibility prevents residual variances from being distorted by unmodeled curvature in the mean structure \citep{jeanselme2025assessing}, a critical prerequisite for reliable variance change detection.  

For $(x_i, y_i), i=1,\dots,n.$ A smoothed function $f(\cdot)$ can be estimated using spline regression such that
\begin{align}
    y_i = f(x_i) + \epsilon_i
\end{align}
where $\epsilon_i, i=1,\dots,n$ denotes the independent random error with variance $\sigma_i^2$. We seek to detect a variance change point $x_{\tau}$ such that $\sigma_1 = \sigma_2 = \dots = \sigma_{\tau} \neq \sigma_{\tau+1} = \dots = \sigma_{n}$.

Penalized splines \citep{marx1996} are particularly useful for variance change point detection because they provide a flexible and smooth estimate of the mean structure. The smoothness is controlled by selecting an appropriate order of B-spline base functions, usually order three is enough \citep{marx2010}. Furthermore, penalized splines prevent overfitting by introducing a regularization parameter $\lambda$. The unknown regression function $f(\cdot)$ is estimated as a linear combination of coefficients and B-spline base functions,
\begin{align}
   f(x_i) = \sum_{j=1}^{K}\beta_j B_j(x_i)
\end{align}
where $K$ is the pre-selected number of B-splines, $\beta_j, j=1,\dots,K$ are the coefficients, and $B_j(\cdot)$ is the $j$th B-spline base function. In this paper, we select K=10 and B-spline base functions with order three to ensure smoothness over the second derivative. The function $f(\cdot)$ can be estimated by optimizing over the following regularized likelihood
\begin{align}
 \sum_{i=1}^{n}[ y_i - \sum_{j=1}^{K}\beta_j B_j(x_i)]^2 + \lambda\sum_{j=3}^{K}(\Delta^2\beta_j)^2
\end{align}
where $\Delta^2(\cdot)$ denotes the second order difference, and $\lambda=0.6$ is a pre-selected penalty parameter which suffices in our case. To select an appropriate $\lambda$ for any other specific data, one can use cross-validation with AIC or BIC \citep{arijit2011}.   

To test the existence of a variance change point, construct the following hypothesis
\begin{align}
 H_o&: \sigma_1 = \dots = \sigma_n  \\
 H_1&: \sigma_1 = \sigma_2 = \dots = \sigma_{\tau} \neq \sigma_{\tau+1} = \dots = \sigma_{n} \notag
\end{align}

Considering the following likelihood \citep{du2018} on a point $\tau = 1, \dots, n$,
\begin{align}
l(\tau) = & \tau \mathrm{log}\{\frac{1}{\tau}\sum_{i=1}^{\tau}[y_i - f(x_{\tau})]^2\} + \\ & (n-\tau)\mathrm{log}\{\frac{1}{n-\tau}\sum_{i=\tau+1}^{n}[y_i-f(x_{\tau})]^2\} \notag
\end{align}

a test statistic $T_n$ can be constructed based on the Schwartz information criterion (SIC) \citep{gideon1978},
\begin{align}
T_n = \mathrm{log}n - \mathrm{min}_{1<\tau<n}\{ l(x_{\tau}) - l(x_{n})\}
\end{align}
Given $\alpha$ as our significance level, we reject $H_0$ if 
\begin{align}
a_n(\mathrm{log}n)^{1/2}T_n - b_n\mathrm{log}(n) > -\mathrm{log}\{-\mathrm{log}(1-\alpha)/2\}
\end{align}
where $ a_n = [2\mathrm{log}(\mathrm{log}(n))]^{1/2}/\mathrm{log}(n)$ and $b_n = \{ 2log(log(n)) + \frac{1}{2}log[log(log(n))] - log(\Gamma(\frac{1}{2}))\}/log(n) $.

To detect all variance change points, we adopt a sequential hypothesis testing framework. The process begins by identifying the first significant change point $x_{\tau_1}$ using the entire dataset. Once $x_{\tau_1}$ is detected, the data is split into two segments: $[x_{1}, x_{\tau_1}]$ and $[x_{\tau_1+1}, x_{n}]$. Hypothesis testing is then applied to each segment independently to detect additional change points, such as $x_{\tau_2}$ and $x_{\tau_3}$, if present. This iterative procedure continues until no further significant change points are found. By progressively narrowing the search intervals, the method avoids overlapping variance regimes and ensures that each structural break is accurately and independently identified.

%% file: sec/5_experiment.tex
\section{Experiments}
\label{experiment}

\subsection{Benchmark evaluation based on PCD}

To evaluate the effectiveness of PCD as a spatial reliability metric, we conducted a comprehensive benchmark study spanning 16 experiments across four environmental conditions (clear daylight, rainy daylight, rainy night, and rainy night under streetlights), two target objects (a vehicle and a dummy pedestrian), and two perception tasks (object detection and instance segmentation). Each perception model was assessed over the full distance range using both aPCD and standard evaluation metrics—AP$_{50:95}$, AP$_{50}$, AR, and F1$_{50}$ \cite{padilla2020survey}.

\noindent \textbf{Model Implementation.} On the SensorRainFall dataset, we ran a series of perception benchmarks over sequential images. Five object detection models were evaluated: Deformable DETR \citep{zhu2021deformable}, Grounding DINO \citep{liu2024groundingdinomarryingdino}, DyHead \citep{DyHead_CVPR2021}, YOLOX \citep{ge2021yolox}, and GLIP \citep{li2022groundedlanguageimagepretraining}. Five instance segmentation models were evaluated: Mask R-CNN \citep{He_2017}, ConvNeXt-V2 \citep{woo2023convnext}, SOLOv2 \citep{wang2020solov2}, Mask2Former \citep{cheng2022masked}, and RTMDet \citep{lyu2022rtmdet}. Among these, GLIP and Grounding DINO are multi-modal models, while the others are deterministic. All selected models have demonstrated state-of-the-art performance on widely used open-source benchmarks such as Microsoft COCO \citep{lin2014microsoft}, Objects365 \citep{Shao2019ICCV}, LVIS \citep{Gupta_2019_CVPR}, iSAID \citep{waqas2019isaid}, and OID \citep{kuznetsova2020open}.

The following configurations are used for each model. Deformable DETR uses a ResNet-50 backbone with a two-stage refinement module to improve region proposal quality. Grounding DINO adopts a Swin-B backbone with high-dimensional embeddings and a modified neck supporting 256/512/1024 features. DyHead integrates an ATSS detector with a Swin-Transformer backbone and an FPN-DyHead neck, pretrained on ImageNet-22K and COCO with multi-scale augmentation. YOLOX employs a CSPDarknet backbone (0.33× depth, 0.5× width), a YOLOXPAFPN neck, and a YOLOX head trained on COCO with heavy augmentation. GLIP incorporates a Swin-S backbone, a modified FPN neck, and an enhanced bounding-box head with early fusion and multiple dynamic head blocks. Mask R-CNN with a ResNeXt-101-64×4d backbone uses an FPN and multi-scale training under a 3× COCO schedule with polynomial LR decay. ConvNeXt-V2 uses a Mask R-CNN framework with a ConvNeXt-V2-B backbone pretrained via FCMAE, paired with an FPN and trained with large-scale jittering under a 3× schedule. SOLOv2 adopts a ResNeXt-101 backbone with deformable convolutions, an FPN neck, and multi-scale training under a 3× schedule. Mask2Former uses a Swin-S backbone with a pixel decoder and transformer decoder head, trained with large-scale jittering for 50 epochs. RTMDet employs a CSPNeXt-X backbone with a P5 design, a CSPNeXtPAFPN neck, and an RTMDetSepBNHead with a DistancePointBBoxCoder. The detailed model configurations are provided in the Supplementary Material.

All perception benchmarks perform object detection and instance segmentation directly on the SensorRainfall dataset without any fine-tuning. For GLIP and Grounding DINO, the text prompt ``car'' is used to guide the object detection task. During inference, no confidence score threshold is applied; instead, each model outputs the bounding box or instance mask labeled ``car'' or ``person'' with the highest confidence score.

The software environment was based on Python 3.11 running on Rocky Linux 9.3. The model was trained on a high-performance GPU workstation with dual Intel Xeon Gold 6442 CPUs @ 2.60 GHz, 512 GB RAM, and one Nvidia Tesla H100 80 GB GPU.

\noindent \textbf{Benchmark comparison.} The full results of all 16 experiments are detailed in the Supplementary Material, while three representative examples are summarized in Table~\ref{RPD example results}.  Example 1 reflects the dominant pattern observed in most experiments (11 out of 16): aPCD effectively captures the maximum distance at which each model maintains reliable detection across varying thresholds of $y^{thres}$ and $p^{thres}$. Models like Mask2Former, which achieve higher aPCD, also perform well on traditional metrics, indicating alignment between spatial reliability and detection quality. 


\begin{table}[htpb]
  \caption{Example evaluation results of perception benchmarks. }
  \label{RPD example results}
  \centering
  \footnotesize
  \makebox[\linewidth][c]{
  \begin{tabular}{lccccc}
    \toprule
    Model & aPCD (m)& AP$_{50:95}$& AP$_{50}$ &AR &F1$_{50}$\\
    \midrule
    \multicolumn{6}{c}{Example 1: Instance segmentation, Vehicle, Clear daylight}\\
    Mask R-CNN\citep{He_2017}& 89.750& 0.376& 0.579& 0.381&0.736\\
    ConvNeXt-V2\citep{woo2023convnext}& 89.454& 0.395& 0.553& 0.399&0.715\\
    SOLOv2\citep{wang2020solov2}& 36.572& 0.233& 0.276& 0.237&0.438\\
    Mask2Former\citep{cheng2022masked}& \textbf{107.095}& \textbf{0.423}& \textbf{0.633}& \textbf{0.427}&\textbf{0.778}\\
    RTMDet\citep{lyu2022rtmdet}& 43.471& 0.349& 0.593& 0.353&0.747\\
    \midrule
    \multicolumn{6}{c}{Example 2: Object detection, Vehicle, Rainy night}\\
    Deformable DETR\citep{zhu2021deformable}& 3.846& 0.056& 0.133& 0.058&0.239\\
    Grounding DINO\citep{liu2024groundingdinomarryingdino}& 29.583& 0.125& 0.297& 0.128&0.461\\
    DyHead\citep{DyHead_CVPR2021}& 21.546& \textbf{0.144}& \textbf{0.362}& \textbf{0.146}&\textbf{0.534}\\
    YOLOX\citep{ge2021yolox}& 23.772& 0.106& 0.212& 0.109&0.353\\
    GLIP\citep{li2022groundedlanguageimagepretraining}& \textbf{37.299}& 0.133& 0.288& 0.136&0.451\\

    \midrule
    \multicolumn{6}{c}{Example 3: Instance segmentation, Pedestrian, Clear daylight}\\
    Mask R-CNN\citep{He_2017}
&21.091& 0.048  
& 0.133 
& 0.050&0.239\\
    ConvNeXt-V2\citep{woo2023convnext}
&23.448& \textbf{0.068}  
& \textbf{0.175}
& \textbf{0.070}&\textbf{0.303}\\
    SOLOv2\citep{wang2020solov2}
& 9.261& 0.040& 0.103
& 0.043&0.191\\
    Mask2Former\citep{cheng2022masked}
& \textbf{23.547}& 0.055  
& 0.154
& 0.057&0.272\\
    RTMDet\citep{lyu2022rtmdet}& 5.941& 0.033  & 0.069& 0.035&0.135\\

    \bottomrule
  \end{tabular}
  }
\end{table}

\begin{figure}
  \centering
  \includegraphics[scale=0.6]{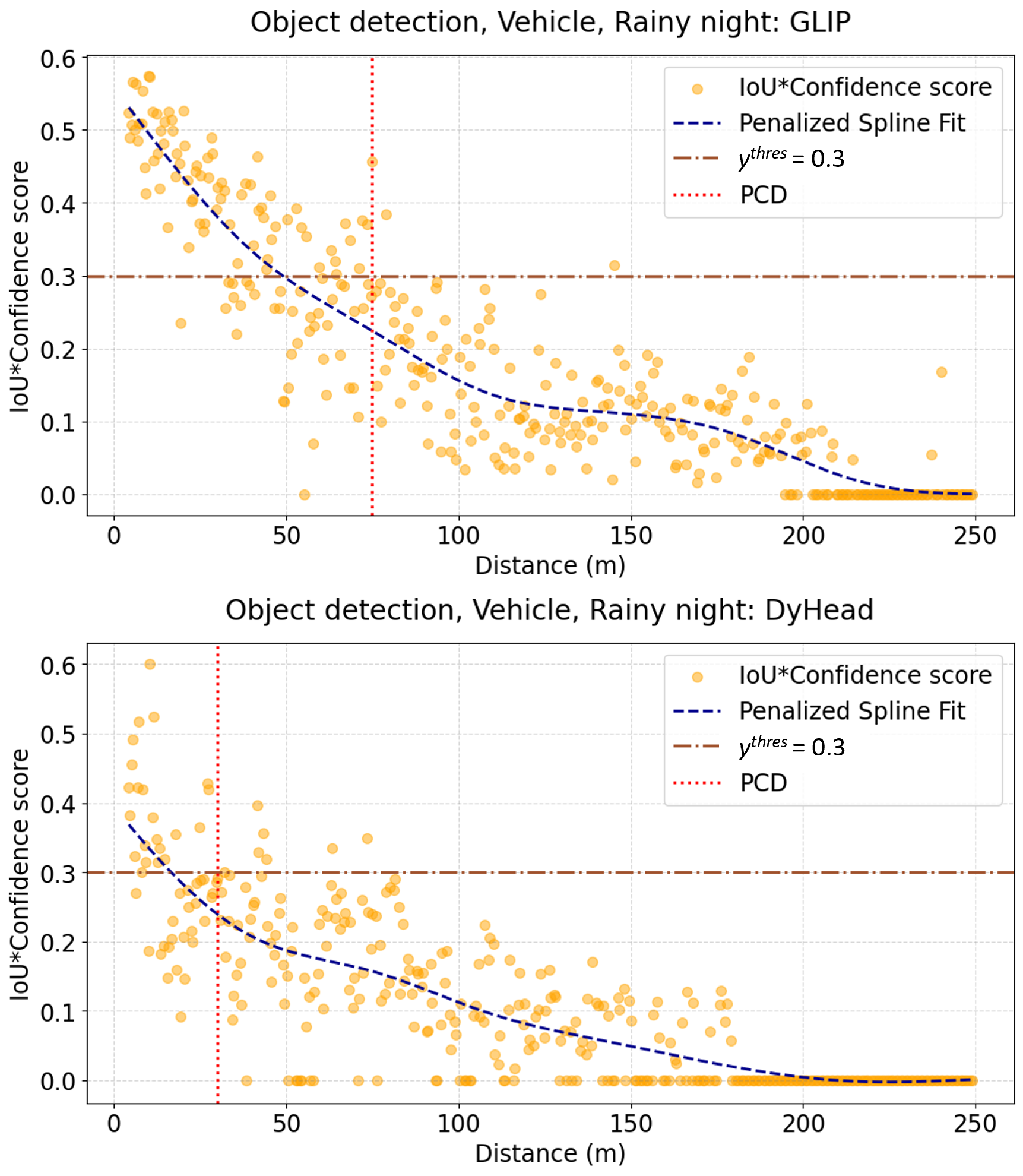}
  \caption{Evaluation results of Example 2: Object detection, Vehicle, Rainy night. GLIP vs. DyHead (PCD threshold: $p^{thres}$ = 0.3 and $y^{thres}$ = 0.3).}
  \label{fig:example 2}
\end{figure}

Example 2 in Table~\ref{RPD example results} demonstrates GLIP achieves the highest aPCD and DyHead attains the highest values for AP$_{50:95}$, AP$_{50}$, AR, and F1$_{50}$. Figure \ref{fig:example 2} provides insight into this discrepancy. Compared with DyHead, GLIP exhibits (1) consistently higher IoU $\times$ Confidence scores, (2) lower variance across distance, and (3) a notably shorter “failure tail,” i.e., a shorter region at long ranges (beyond 200m) where the model failed to detect the object. This indicates that GLIP maintains stable outputs over DyHead. Metrics like AP or AR aggregate performance across IoU thresholds and confidence-based ranking but are largely insensitive to how performance degrades with distance. In contrast, aPCD directly quantifies where the model begins to fail in a physically meaningful way. aPCD metric highlights this reliability advantage, revealing performance characteristics that conventional metrics overlook.

\begin{figure}
  \centering
  \includegraphics[scale=0.6]{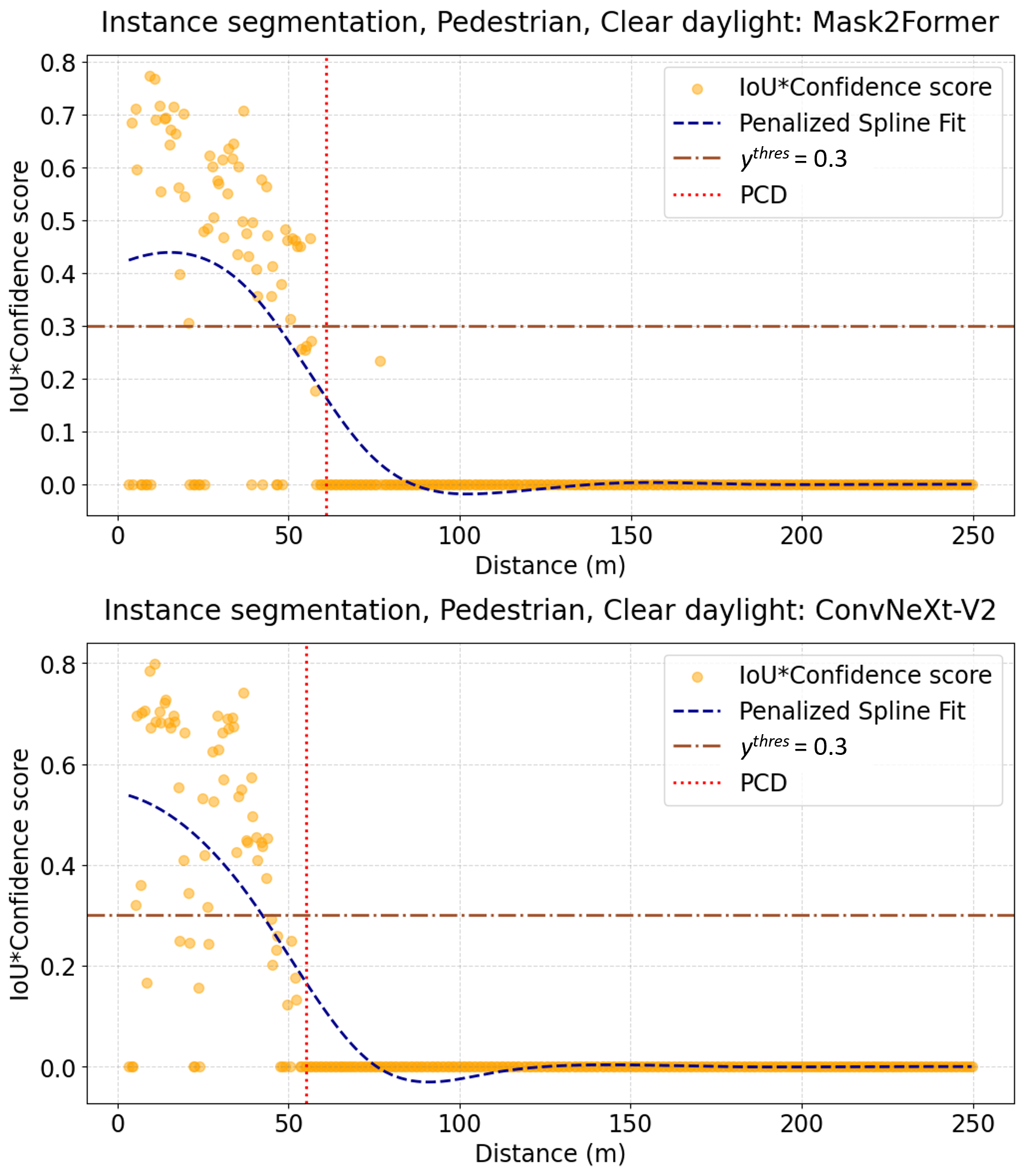}
  \caption{Evaluation results of Example 3: Instance segmentation, Pedestrian, Clear daylight. Mask2Former vs. ConvNeXt-V2 (PCD threshold: $p^{thres}$ = 0.3 and $y^{thres}$ = 0.3).}
  \label{fig:example 3}
\end{figure}

Example 3 in Table~\ref{RPD example results} reveals that Mask2Former achieves the highest aPCD, while ConvNeXt-V2 obtains the highest AP$_{50:95}$, AP$_{50}$, AR, and F1$_{50}$. Figure~\ref{fig:example 3} explains this difference. Compared to ConvNeXt-V2, Mask2Former produces consistently higher IoU $\times$ confidence scores at short ranges ($<50$m) and displays lower variance in this region. This means Mask2Former behaves more predictably and provides more stable responses. In contrast, ConvNeXt-V2 shows larger fluctuations and a faster decline toward zero, despite achieving higher AP-based scores. As a result, Mask2Former demonstrates stronger distance-dependent reliability, which is captured directly by the aPCD but not by conventional metrics. The example highlights how aPCD uncovers robustness characteristics that standard metrics overlook, providing a complementary perspective on model performance.

\subsection{Experimental insights}

The key advantage of PCD over conventional evaluation metrics lies in its ability to incorporate perception spatial uncertainty through two adjustable thresholds,  $p^{thres}$ and $y^{thres}$. By varying these thresholds, our experiments demonstrate that PCD not only enables the determination of a safety envelope, but also reveals characteristic behaviors of perception systems under different conditions and serves as a flexible evaluation metric adaptable to diverse operational needs. 

\begin{figure}[htpb]
  \centering
  \includegraphics[scale=0.35]{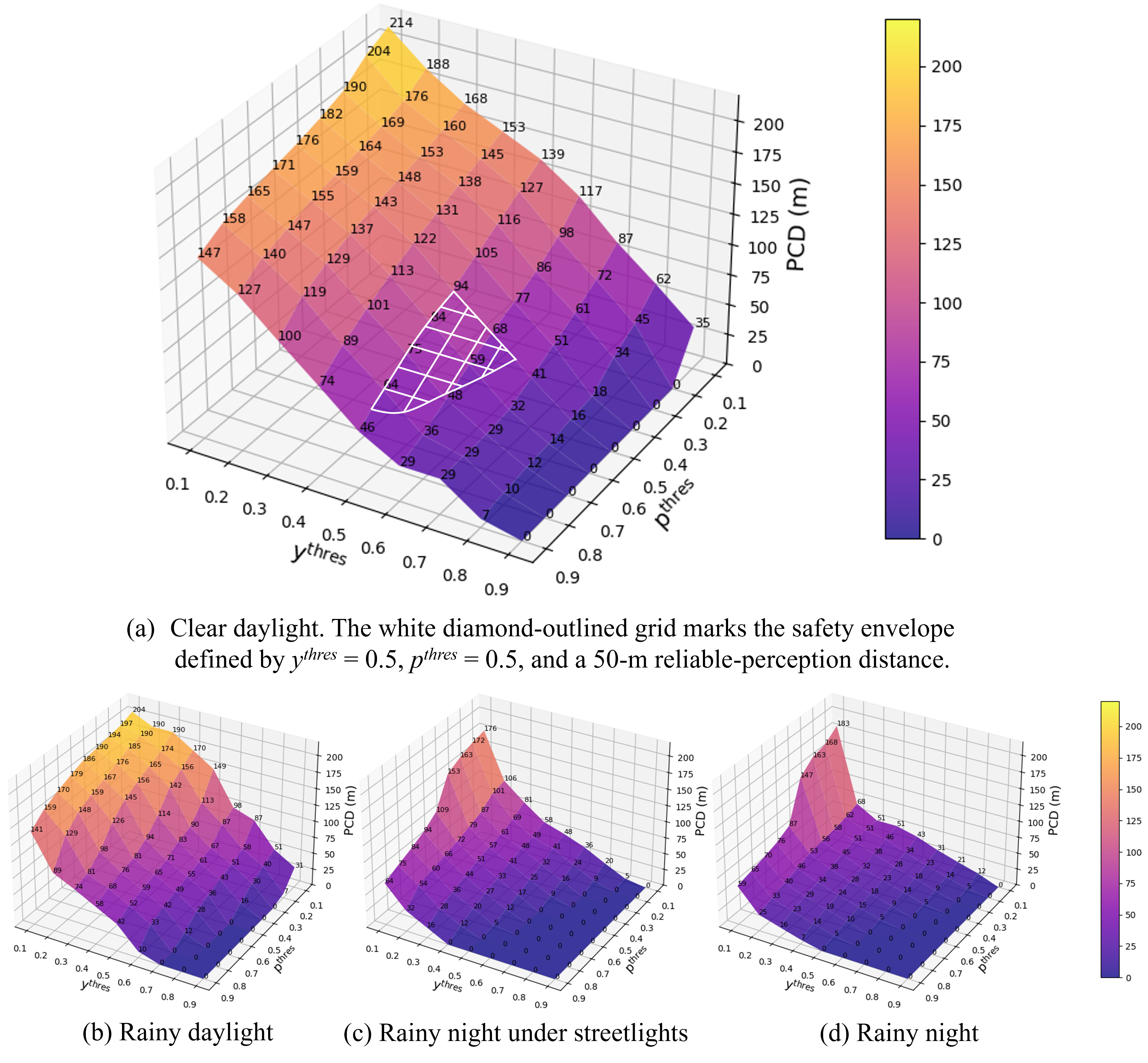}
  \caption{PCD across threshold combinations ($p^{thres}$, $y^{thres}$) and environmental conditions (Object detection, Vehicle: Grounding DINO).}
  \label{fig:various-thresholds}
\end{figure}

\noindent \textbf{Determine the Safety Envelope.} 
PCD, defined over combinations of $p^{thres}$ and $y^{thres}$, provides a practical means to determine the safety envelope of a perception system. The safety envelope represents the operational region within which a perception model can reliably detect objects without risking perception failures. Figure~\ref{fig:various-thresholds}(a) illustrates an example safety envelope for the object detection task under Clear daylight, derived using $p^{thres}$ = 0.5, $y^{thres}$ = 0.5, and a 50-m reliable perception distance. Within this envelope, the ADS can trust the perception outputs for downstream planning, prediction, and control. As expected, the safety envelope identified by PCD changes with environmental conditions and tends to shrink as visibility deteriorates, reflecting reduced perception reliability under severe environments. 

\noindent \textbf{Uncover Perception System Characteristics.}
PCD reveals how perception performance degrades as environmental conditions change. As shown in Figure~\ref{fig:various-thresholds}(a) and (b), PCD varies smoothly under clear and rainy daylight, reaching higher values when both $p^{thres}$ and $y^{thres}$ are low. In contrast, in Figure~\ref{fig:various-thresholds}(c) and (d), rainy night and rainy night under streetlight conditions produce abrupt PCD fluctuations, indicating that the perception system becomes more sensitive to threshold changes and less stable under adverse environments.

\noindent \textbf{Serve as a Flexible Metric.} 
By adjusting $p^{thres}$ and $y^{thres}$ according to operational requirements, practitioners can tailor PCD to emphasize either early detection or high reliability depending on the task, target object, and environmental condition. Lower thresholds ($p^{thres}, y^{thres} \approx 0.1$–$0.3$) extend the detection range but allow less reliable predictions—useful in safety-critical situations such as emergency braking for vulnerable road users, where detecting an object early (even with low confidence) is more important than strict precision. Higher thresholds ($p^{thres}, y^{thres} \approx 0.6$–$0.9$) enforce strict reliability at the cost of shorter detection distances; for example, high-confidence perception is necessary for tasks like high-speed lane keeping or precise maneuvering, where false positives can destabilize planning. For most real-world scenarios, mid-range thresholds ($p^{thres}, y^{thres} \approx 0.4$–$0.6$) offer a practical balance by providing reasonable detection range while maintaining stable and trustworthy predictions—such as in typical urban driving, where moderate confidence is sufficient for safe planning. 



\subsection{\texorpdfstring{Effectiveness of variance change point detection}{Effectiveness of variance change point detection}}

Variance change point detection is an integral part of the PCD calculation, allowing the variance within each segment to be estimated accurately. Because the procedure is based on hypothesis testing, the statistical significance of the decision depends on both sample size and effect size \cite{wasserman2013all}. This section presents two simulation studies designed to evaluate how these two factors influence the performance of the change point detection method.

\noindent \textbf{Impact of Sample Size.} 
To assess how sample size affects the statistical power of the variance change point detection method, we conduct a controlled simulation using synthetic 1D datasets with predefined variance shifts. Each dataset follows a decreasing linear trend with Gaussian noise, and each change point enforces a clear effect size: variance increases by 5–10× or decreases to 0.1–0.2× of its previous value. For each combination of sample size n and true number of change points k, we generate 1,000 replications and apply the proposed method. By summarizing the average number of detected change points and its variability, the experiment evaluates detection accuracy and robustness across sample sizes while holding effect size fixed.

\begin{figure}
  \centering
  \includegraphics[scale=0.27]{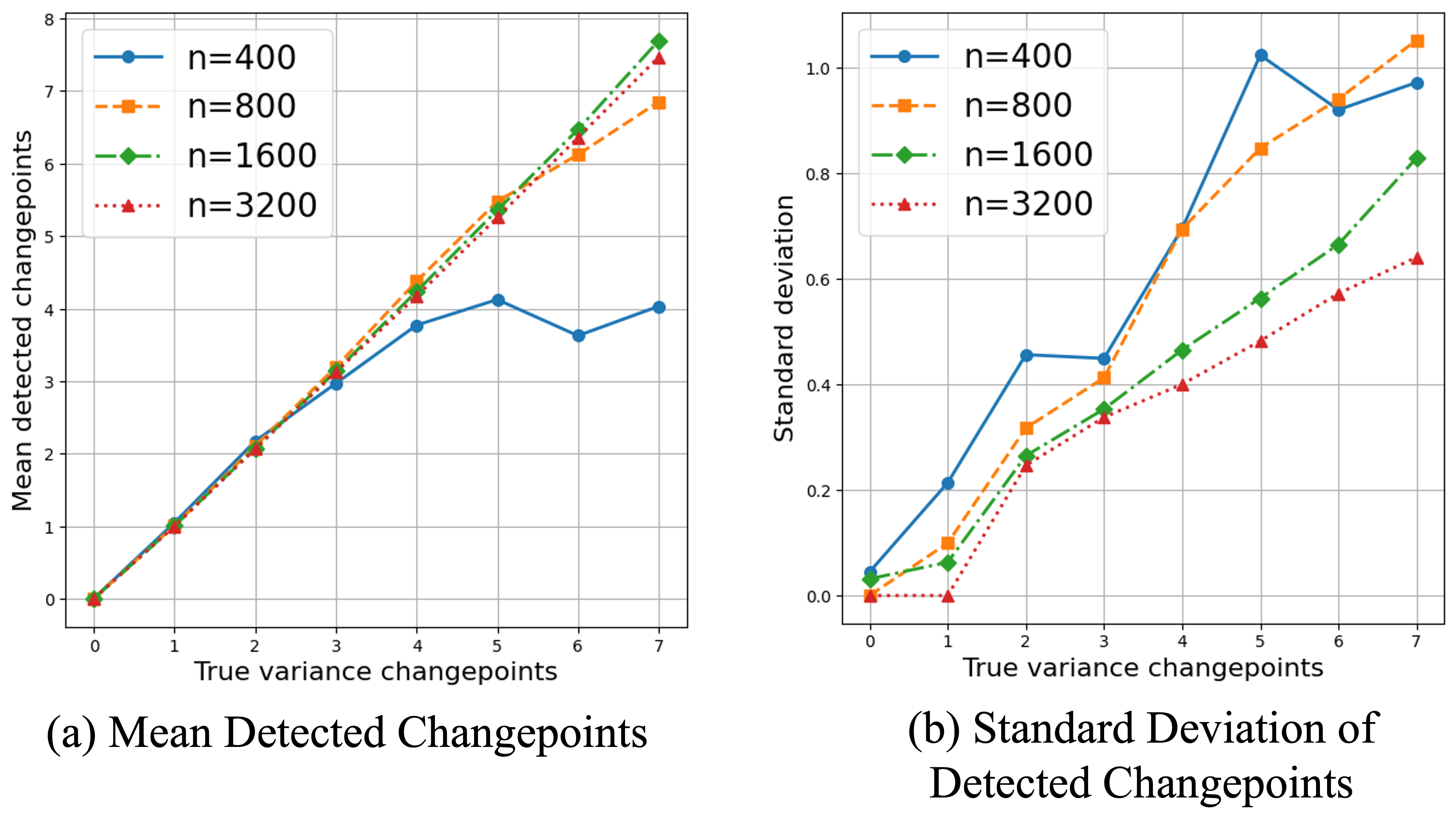}
  \caption{Mean and standard deviation of detected variance change points across different sample sizes}
  \label{fig:sim_mean_var}
\end{figure}

The results of this simulation are presented in Figure~\ref{fig:sim_mean_var}. Both panels show a clear trend in which larger sample sizes lead to more accurate and more stable variance change point detection. The mean number of detected change points closely follows the true number when the number of change points is fewer than four. The variability of detected change points is also small in this range. This level of accuracy and stability is sufficient for the PCD calculation scenario, where segment-level variance estimates require only moderate segment lengths to remain robust.

\begin{figure}
  \centering
  \includegraphics[scale=0.25]{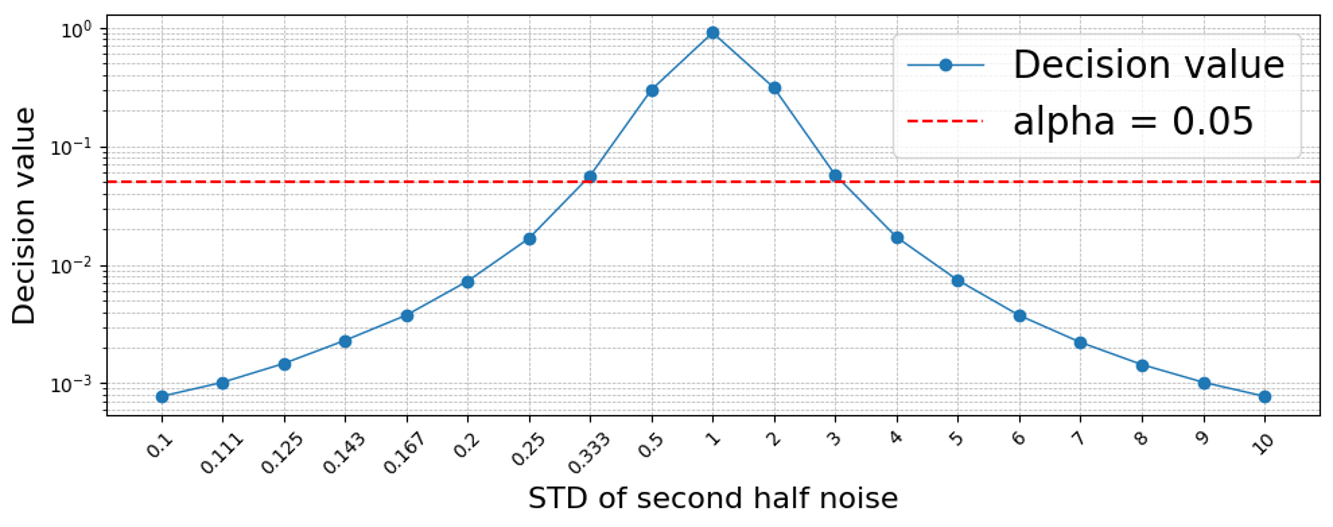}
  \caption{Decision value as a function of variance ratio under 50--50 samples.}
  \label{fig:eff_dec}
\end{figure}

\noindent \textbf{Impacts of Effect Size.} 
The effect size refers to the difference in variance among groups. From Equation~(7), the test statistic is given by $S = a_n(\log n)^{1/2}T_n - b_n\log(n)$, and a change point is detected whenever $1 - \exp(-2e^{-S}) < \alpha$ according to the decision rule. In this experiment, we construct multiple datasets, each containing a single change point at the midpoint. The left segment is identical across datasets and follows a decreasing linear trend with Gaussian noise $N(0,1)$, while the right segment varies across datasets, taking noise distributions ranging from $N(0,0.01)$ to $N(0,100)$, with each dataset using one fixed variance. As shown in Figure~\ref{fig:eff_dec}, the detection statistic becomes increasingly significant as the variance difference grows, illustrating ratio under 50--50 samples, showing that a variance increase of approximately $3\times$ is sufficient for significant detection for change points.

%% file: sec/6_conclusion.tex
\section{Conclusion}
\label{conclusion}

This study proposes PCD, a novel metric that quantifies the maximum range at which a perception system maintains consistent confidence above a specified threshold. PCD estimates both the mean and the variance of IoU $\times$ Confidence score as a function of distance. It reveals that perception performance exhibits heteroscedastic variance, exposing limitations of traditional static metrics. To support the development and evaluation of PCD, we introduce the SensorRainFall dataset, a controlled collection that captures perception performance across four environmental conditions: clear daylight, rainy daylight, rainy night under streetlights, and rainy night. Through extensive experiments on various state-of-the-art models, we demonstrate that PCD provides a more robust and informative assessment of perception stability under dynamic real-world conditions. Although this study focuses on object detection and instance segmentation within SensorRainFall dataset, future work should extend PCD to additional datasets and a broader range of perception tasks. By improving the reliability evaluation of perception systems, PCD has the potential to enhance autonomous-driving safety, while the broader societal impacts of full autonomy.


%% file: sec/X_suppl.tex
\clearpage
\setcounter{page}{1}
\maketitlesupplementary




\section{SensorRainFall Dataset Description}


\begin{figure}
  \centering
  \includegraphics[scale=0.4]{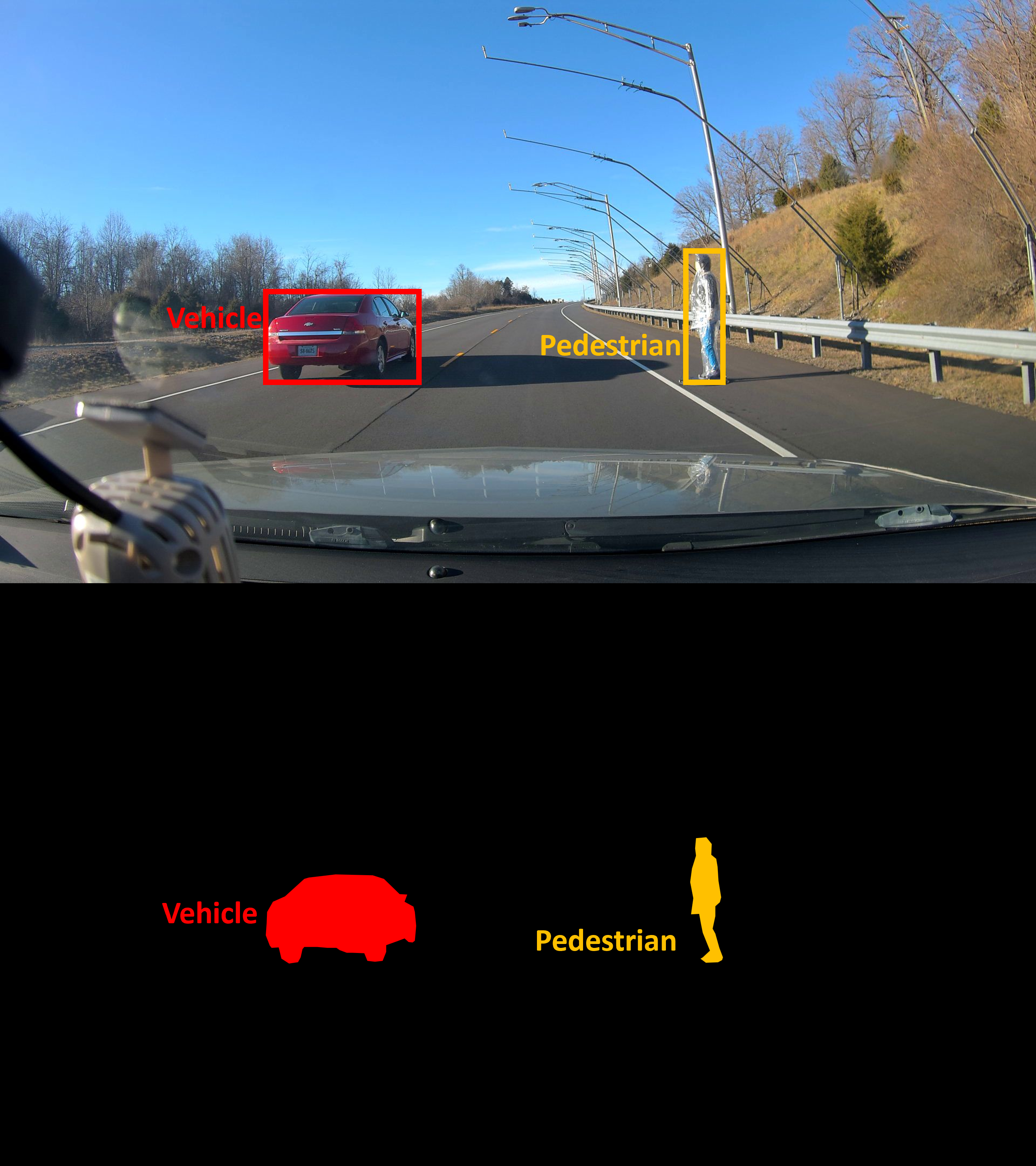}
  \caption{Example of ground-truth bounding box and instance mask under clear daylight (vehicle: 9.751 m, pedestrian: 9.817 m from ego vehicle).}
  \label{fig:bounding box+mask}
\end{figure}

The SensorRainFall dataset offers a uniquely high-fidelity benchmark for evaluating perception robustness under precipitation. This dataset enables modeling of perception variability across distance, rainfall intensity, and lighting conditions, making it particularly well suited for uncertainty-aware metrics such as PCD.

The SensorRainFall dataset was developed under realistic yet highly controlled conditions. It was collected on the Virginia Smart Roads facility, which supports precise simulation of weather and lighting scenarios. The rainfall was generated using weather simulation towers spaced at 10 meters, producing two nominal rainfall intensities—20 mm/h (heavy rain) and 40 mm/h (excessive rain)—as verified using a Weather Characterizer system based on OTT Parsivel2 laser disdrometry. 

To ensure safety and repeatability, target objects were positioned one lane offset from the ego vehicle’s path—an approach validated during pilot testing to produce comparable sensor responses to in-lane positioning. Testing was conducted across three ambient lighting settings (day, night, and night with overhead 3000K LED streetlights) and four sensor modalities (camera, lidar, radar, and an ADAS-specific object detection camera), with vehicle speed ranging between 10–55 mph. 

Target objects consisted of a red sedan and an ISO 19206-compliant surrogate pedestrian, with the latter wrapped in a plastic poncho to prevent rain damage. Pilot testing confirmed that the poncho created a “halo” effect that improved visibility; thus, it was consistently applied across both rainy and non-rainy runs to isolate rainfall effects. Target positions were recorded using differential GPS (DGPS), and all sensor outputs were time-synchronized using a Robot Operating System (ROS)–based architecture. A summary of the SensorRainFall dataset, along with descriptive statistics of the distance between the ego vehicle and target objects across all environmental conditions, is provided in Table \ref{dataset description}.

\begin{table}[htpb]
  \caption{SensorRainfall dataset overview.}
  \label{dataset description}
  \centering
  \makebox[\linewidth][c]{
  \begin{tabular}{>{\raggedright\arraybackslash}p{1.3cm}>{\centering\arraybackslash}p{1.2cm}>{\centering\arraybackslash}p{1.2cm}>{\centering\arraybackslash}p{1cm}>{\centering\arraybackslash}p{1.8cm}}
    \toprule
     & Clear daylight & Rainy daylight & Rainy night & Rainy night under streetlights \\
    \midrule
     \multicolumn{5}{c}{\textbf{Target: Vehicle}}\\
     Count of images& 278 & 317 & 354 & 281 \\
     Mean*
& 107.989 & 122.251 & 119.798 & 107.761 \\
     Std*
& 61.852  & 69.800  & 72.988  & 61.088  \\
     Min*
& 5.553   & 5.476   & 4.412   & 4.888   \\
     Median*
& 105.976 & 121.208 & 114.734 & 107.240 \\
     Max*& 214.247 & 240.669 & 248.937 & 214.637 \\
    \midrule
     \multicolumn{5}{c}{\textbf{Target: Pedestrian}}\\
     Count of images& 279 & 230 & 352 & 280 \\
     Mean*
& 107.661 & 152.615 & 120.919 & 108.593 \\
     Std*
& 61.947  & 55.303  & 72.654  & 60.889  \\
     Min*
& 4.442   & 4.596   & 4.639   & 4.501   \\
     Median*
& 105.847 & 156.005 & 115.997 & 107.946 \\
     Max*& 214.081 & 240.042 & 249.275 & 215.011 \\
    \bottomrule
  \end{tabular}
  }
  \vspace{2pt}
  \begin{minipage}{\linewidth}
  \footnotesize\raggedright *\textit{Descriptive statistics of the distance (m) between the ego vehicle and the target.}
  \end{minipage}
\end{table}

Each image in the dataset was carefully examined by trained researchers and manually annotated with ground-truth bounding boxes and pixel-level segmentation masks for both the vehicle and the pedestrian, as illustrated in Figure \ref{fig:bounding box+mask}. The annotations include precise bounding box coordinates (four corners) and detailed instance masks that capture the full spatial extent of each object. These high-resolution annotations enable both object detection and instance segmentation tasks and provide reliable ground truth for evaluating spatial accuracy across perception models.




\section{Complete benchmark results}

The complete results of all 16 experiments are presented in this section. Tables~\ref{object detection_car_num of change point}, ~\ref{object detection_person_num of change point}, ~\ref{segmentation_car_num of change point}, and ~\ref{segmentation_person_num of change point} report the detected variance change points at $\alpha = 0.05$ for all benchmarks, including both the count and corresponding distance locations (m). Tables~\ref{eval results_Object detection_car}, ~\ref{eval results_Object detection_person}, ~\ref{eval results_segmentation_car}, and ~\ref{eval results_segmentation_person} provide comprehensive evaluation results using the proposed aPCD metric and conventional metrics (AP$_{50:95}$, AP$_{50}$, AP$_{75}$, AP$_S$, AP$_M$, AP$_L$, AR, and F1$_{50}$).


\begin{table*}[htpb]
\caption{Detected variance change points ($\alpha=0.05$) across benchmarks: Object detection, Vehicle.}
  \label{object detection_car_num of change point}
  \centering
  \footnotesize
  \makebox[\linewidth][c]{
  \begin{tabular}{lcccccc}
    \toprule
    Model & Number of variance change points & $x_{\tau_{1}}$ (m) & $x_{\tau_{2}}$ (m) & $x_{\tau_{3}}$ (m) & $x_{\tau_{4}}$ (m) &$x_{\tau_{5}}$ (m)\\
    \midrule
    \multicolumn{7}{c}{Clear daylight} \\
    Deformable DETR & 0 & – & – & – & – & – \\
    Grounding DINO & 2 & 29.407 & 154.556 & – & – & –\\
    DyHead & 2 & 109.377 & 198.193 & – & –& – \\
    YOLOX & \textbf{3} & 36.513 & 133.427 & 170.522 & –& – \\
    GLIP & 2 & 81.399 & 154.556 & – & –& – \\
    \midrule
    \multicolumn{7}{c}{Rainy daylight} \\
    Deformable DETR & 3& 130.054 & 189.140 & 198.106 & –& – \\
    Grounding DINO & 4& 81.681 & 87.833 & 190.611 & 217.114& – \\
    DyHead & 4& 16.037 & 84.795 & 195.101 & 208.407 & –\\
    YOLOX & 3 & 119.157 & 138.339 & 153.980 & –& – \\
 GLIP & 3& 41.514 & 110.335 & 158.641 &–& – \\
    \midrule
 \multicolumn{7}{c}{Rainy night}\\
 Deformable DETR 
& 5 & 31.169& 51.627& 122.244 &211.178& 218.465\\
 Grounding DINO 
& 4& 51.627& 164.393& 209.696& 215.572& –\\
 DyHead 
& 3& 82.943& 117.468& 211.178& –& –\\
 YOLOX 
& 4 & 70.742& 87.990& 104.899& 186.902& – \\
 GLIP & 2& 148.341& 240.375&– & –& –\\
 \midrule
    \multicolumn{7}{c}{Rainy night under streetlights}\\
    Deformable DETR 
& 3& 46.315& 54.084& 146.579& –& – \\
    Grounding DINO 
& 3& 49.268& 107.240& 177.282& – & –\\
    DyHead 
& 3& 22.784& 39.309& 148.470& –& –\\
 YOLOX 
&3 & 65.933&77.431 &165.402 & –& –\\
    GLIP & 2& 92.664& 142.363& – & –& – \\
    \bottomrule
  \end{tabular}
  }
\end{table*}

\begin{table*}[htpb]
\caption{Detected variance change points ($\alpha=0.05$) across benchmarks: Object detection, Pedestrian.}
\label{object detection_person_num of change point}
\centering
\footnotesize
\makebox[\linewidth][c]{
\begin{tabular}{lcccc}
\toprule
Model & Number of variance change points & $x_{\tau_{1}}$ (m) & $x_{\tau_{2}}$ (m) & $x_{\tau_{3}}$ (m) \\
\midrule
\multicolumn{5}{c}{Clear daylight} \\
Deformable DETR & 3 & 60.610 &  133.429& 170.443 \\
Grounding DINO & 2 &57.731& 109.034 &–\\
DyHead & 3 & 81.144 & 109.438 & 189.473 \\
YOLOX &2 &66.391& 125.249& – \\
GLIP & 2 & 67.841 & 116.894 & – \\
\midrule

\multicolumn{5}{c}{Rainy daylight} \\
Deformable DETR & 3 & 71.048 & 137.862 & 173.446 \\
Grounding DINO & 2 & 60.182 & 159.687 & – \\
DyHead & 3 & 66.157 & 111.520 & 168.912 \\
YOLOX & 3 & 42.607 & 128.127 & 182.582 \\
GLIP & 3 & 107.546 & 142.961 & 171.961 \\
\midrule

\multicolumn{5}{c}{Rainy night} \\
Deformable DETR & 2 & 49.710 & 105.067 & – \\
Grounding DINO & 2 & 66.170 & 109.308 & – \\
DyHead & 2 & 76.553 & 109.308 & – \\
YOLOX & 2 & 57.243 & 87.212 & – \\
GLIP & 2 & 76.552 & 118.000 & – \\
\midrule

\multicolumn{5}{c}{Rainy night under streetlights} \\
Deformable DETR & 2 & 83.928 & 134.777 &  – \\
Grounding DINO & 2 & 57.603 & 121.279 & – \\
DyHead & 2 & 40.958 & 122.783 &  – \\
YOLOX & 1 & 77.980 & – &  – \\
GLIP & 2 & 64.637 & 124.293 &  – \\
\bottomrule
\end{tabular}
}
\end{table*}

\begin{table*}[htpb]
\caption{Detected variance change points ($\alpha=0.05$) across benchmarks: Instance segmentation, Vehicle.}
  \label{segmentation_car_num of change point}
  \centering
  \footnotesize
  \makebox[\linewidth][c]{
  \begin{tabular}{lccccc}
    \toprule
    Model & Number of variance change points & $x_{\tau_{1}}$ (m) & $x_{\tau_{2}}$ (m) & $x_{\tau_{3}}$ (m) & $x_{\tau_{4}}$ (m) \\
    \midrule
    \multicolumn{6}{c}{Clear daylight} \\
    Mask R-CNN
& 3& 33.670& 60.446& 152.926& –\\
    ConvNeXt-V2
& 3& 75.422& 131.801& 175.657& –\\
    SOLOv2
& 2& 72.465& 121.949& –& –\\
    Mask2Former
& 4& 54.697& 69.518& 167.761& 173.679\\
    RTMDet& 3& 46.122& 150.112& 186.139& –\\
    \midrule
    \multicolumn{6}{c}{Rainy daylight} \\
    Mask R-CNN& 2& 141.458& 159.004& –& –\\
    ConvNeXt-V2& 2& 63.918& 153.980& –& –\\
    SOLOv2& 3& 52.074& 141.458& 163.257& –\\
    Mask2Former& 4& 41.514& 84.795& 187.994& 202.521\\
 RTMDet& 4& 16.037& 122.468& 158.641&181.621\\
    \midrule
 \multicolumn{6}{c}{Rainy night}\\
 Mask R-CNN& 4& 70.742& 141.67& 178.982& 199.641\\
 ConvNeXt-V2& 3&17.985& 88.327& 186.902& –\\
 SOLOv2& 3& 60.461 & 193.126& 213.022 & –\\
 Mask2Former& 3& 62.057& 108.764& 200.74 &–\\
 RTMDet& 4& 50.379& 97.558& 191.583& 210.083 \\
 \midrule
    \multicolumn{6}{c}{Rainy night under streetlights}\\
    Mask R-CNN& 2&95.259& 157.346& –& –\\
    ConvNeXt-V2& 3& 91.167& 107.612& 186.633&– \\
    SOLOv2& 3&56.666& 74.449& 174.194 &– \\
 Mask2Former&2 & 68.520& 165.020 &–&–\\
    RTMDet& 3& 95.259& 194.742& 211.892&–\\
    \bottomrule
  \end{tabular}
  }
\end{table*}

\begin{table*}[htpb]
\caption{Detected variance change points ($\alpha=0.05$) across benchmarks: Instance segmentation, Pedestrian.}
  \label{segmentation_person_num of change point}
  \centering
  \footnotesize
  \makebox[\linewidth][c]{
  \begin{tabular}{lcccc}
    \toprule
    Model & Number of variance change points & $x_{\tau_{1}}$ (m) & $x_{\tau_{2}}$ (m) & $x_{\tau_{3}}$ (m) \\
    \midrule
    \multicolumn{5}{c}{Clear daylight}\\
    Mask R-CNN
& 2& 84.149& 118.716& –\\
    ConvNeXt-V2
& 2&60.610 & 116.894&– \\
    SOLOv2
& 2& 53.436& 107.423& –\\
    Mask2Former
& 2& 77.037& 130.156& –\\
    RTMDet
& 3& 56.314& 104.262&172.027 \\
    \midrule
    \multicolumn{5}{c}{Rainy daylight}\\
    Mask R-CNN& 2  & 92.918& 158.135& –\\
    ConvNeXt-V2&2 & 73.693& 136.679&– \\
    SOLOv2& 2&87.209 &150.370 & –\\
    Mask2Former& 3&79.714 &113.804 & 158.135\\
 RTMDet& 2&66.157 & 113.449& –\\
    \midrule
 \multicolumn{5}{c}{Rainy night}\\
 Mask R-CNN& 0&–&–&– \\
 ConvNeXt-V2&0&–&–&– \\
 SOLOv2&0&–&–&– \\
 Mask2Former&0&–&–&– \\
 RTMDet& 0&–&–&– \\
 \midrule
    \multicolumn{5}{c}{Rainy night under streetlights}\\
    Mask R-CNN& 0&–&–&– \\
    ConvNeXt-V2& 0&–&–&– \\
    SOLOv2& 0&–&–&– \\
 Mask2Former&0&–&–&– \\
    RTMDet& 0&–&–&– \\
    \bottomrule
  \end{tabular}
  }
\end{table*}


\begin{table*}[htpb]
  \caption{Evaluation results of benchmarks: Object detection, Vehicle.}
  \label{eval results_Object detection_car}
  \centering
  \footnotesize
  \makebox[\linewidth][c]{
  \begin{tabular}{lcccccccccc}
    \toprule
    Model & aPCD (m)& Mean of IoU $\times$ Confidence score & AP$_{50:95}$& AP$_{50}$ & AP$_{75}$ & AP$_{S}$ & AP$_{M}$ & AP$_{L}$  &AR &F1$_{50}$\\
    \midrule
    \multicolumn{11}{c}{Clear daylight} \\
    Deformable DETR & \textbf{104.855}& \textbf{0.487} & 0.511& \textbf{0.947}& 0.439& \textbf{0.921}& \textbf{0.982}& \textbf{0.956}& 0.552&\textbf{0.974}\\
    Grounding DINO & 92.685& 0.425 & \textbf{0.553}& 0.777& 0.594& 0.684& \textbf{0.982}& \textbf{0.956}& \textbf{0.568}&0.867\\
    DyHead & 70.990& 0.337 & 0.533& 0.719& \textbf{0.616}& 0.585& \textbf{0.982}& \textbf{0.956}& 0.564&0.832\\
    YOLOX & 73.105& 0.336 & 0.401& 0.548& 0.427& 0.257& \textbf{0.982}& \textbf{0.956}& 0.452&0.716\\
    GLIP & 60.164& 0.257 & 0.421& 0.517& 0.469& 0.310& \textbf{0.982}& \textbf{0.956}& 0.442&0.655\\
    \midrule
    \multicolumn{11}{c}{Rainy daylight} \\
    Deformable DETR & 57.224 & 0.249 & 0.367 & 0.583 & 0.372 & 0.491 & 0.777 & 0.846  & 0.370&0.739
\\
    Grounding DINO & \textbf{86.593} & \textbf{0.354} & \textbf{0.484} & \textbf{0.684} & \textbf{0.526} & \textbf{0.565} & \textbf{0.984} & \textbf{0.923}  & \textbf{0.487}&\textbf{0.814}
\\
    DyHead & 66.010 & 0.274 & 0.468 & 0.634 & 0.533 & 0.500 & \textbf{0.984} & 0.884  & 0.471&0.778
\\
    YOLOX & 52.426 & 0.208 & 0.323 & 0.413 & 0.353 & 0.223 & 0.857 & \textbf{0.923}  & 0.326&0.587
\\
    GLIP & 42.591 & 0.166 & 0.288 & 0.328 & 0.324 & 0.083 & 0.936 & \textbf{0.923}  & 0.292&0.497\\
    \midrule

    \multicolumn{11}{c}{Rainy night} \\
    Deformable DETR & 3.846& 0.021& 0.056 & 0.133& 0.04& 0.029& 0.220& 0.645 & 0.058 &0.239\\
    Grounding DINO & 29.583& 0.156& 0.125& 0.297& 0.105& 0.087& 0.634& \textbf{0.968}&0.128 &0.461\\
    DyHead & 21.546& 0.121& \textbf{0.144}& \textbf{0.362}& 0.107& \textbf{0.154}& \textbf{0.720}& \textbf{0.968}& \textbf{0.146}& \textbf{0.534}\\
    YOLOX & 23.772& 0.102& 0.106& 0.212& 0.107& 0.017& 0.476& \textbf{0.968}& 0.109&0.353\\
    GLIP & \textbf{37.299}& \textbf{0.182}& 0.133& 0.288& \textbf{0.119}& 0.050& 0.707& \textbf{0.968}& 0.136&0.451\\

    \midrule

    \multicolumn{11}{c}{Rainy night under streetlights} \\
    Deformable DETR &4.829 & 0.032& 0.049& 0.121& 0.032& 0.026&0.197 &0.560  &0.052 &0.222\\
    Grounding DINO &\textbf{34.287} & \textbf{0.182}& \textbf{0.235}& 0.505& 0.185& 0.311&\textbf{0.864}& \textbf{0.960}& \textbf{0.238}&0.675\\
    DyHead & 37.115& 0.180& 0.226& \textbf{0.527}& \textbf{0.185}& \textbf{0.379}& 0.848& 0.720 & 0.23& \textbf{0.693}\\
    YOLOX & 32.213& 0.157& 0.187& 0.391& 0.157& 0.195& 0.712& \textbf{0.960}& 0.19 &0.566\\
    GLIP & 34.283&0.145& 0.150& 0.327& 0.128& 0.058& 0.848& 0.920 & 0.154&0.497\\

    \bottomrule
  \end{tabular}
  }
\end{table*}

\begin{table*}[htpb]
\caption{Evaluation results of benchmarks: Object detection, Pedestrian.}
\label{eval results_Object detection_person}
\centering
\footnotesize
\makebox[\linewidth][c]{
\begin{tabular}{lcccccccccc}
\toprule
Model & aPCD (m) & Mean of IoU $\times$ Confidence score & AP$_{50:95}$ & AP$_{50}$ & AP$_{75}$ & AP$_{S}$ & AP$_{M}$ & AP$_{L}$&AR &F1$_{50}$\\
\midrule
\multicolumn{11}{c}{Clear daylight} \\
Deformable DETR & 25.308 & \textbf{0.132}& \textbf{0.135}& \textbf{0.373}& \textbf{0.065}& \textbf{0.222}& 0.875 & 0.800 & \textbf{0.138}&  \textbf{0.547}\\
Grounding DINO & 22.430 & 0.095 & 0.071 & 0.215 & 0.022 & 0.028 & 0.813 &0.867 &0.074& 0.359\\
DyHead & 22.407 & 0.110 & 0.093 & 0.308 & 0.032 & 0.130 & 0.917 & 0.800 &0.096& 0.475\\
YOLOX &13.534 & 0.049& 0.035 & 0.107& 0.010& 0.0 & 0.312 & \textbf{0.933} & 0.038 &0.200\\
GLIP & \textbf{27.392}& \textbf{0.132}& 0.099 & 0.269 & 0.036 & 0.065 & \textbf{0.979}& 0.800 &0.102&0.428\\
\midrule

\multicolumn{11}{c}{Rainy daylight} \\
Deformable DETR &18.724 & 0.094& 0.085 &0.245&0.041 & 0.164& 0.558& 0.474&0.087&0.394\\
Grounding DINO     & 12.596&0.063 &  0.103 &0.242&0.079 &0.129 &0.605& 0.842& 0.106& 0.394\\
DyHead     &20.846 &0.098 & \textbf{0.147}&\textbf{0.336}&0.113& \textbf{0.195} & 0.907 & 0.842 &\textbf{0.149}&\textbf{0.507}\\
YOLOX & 7.991 & 0.029 & 0.042 & 0.101 & 0.028 & 0.039 & 0.116 & 0.789 &0.044& 0.188\\
GLIP     & \textbf{34.512}& \textbf{0.144}& 0.137  &0.302& \textbf{0.116}& 0.145&\textbf{0.930}& \textbf{0.895} & 0.139& 0.467\\
\midrule

\multicolumn{11}{c}{Rainy night} \\
Deformable DETR & 4.563 & 0.022 & 0.043 & 0.077 & 0.043 & 0.0 & 0.488 & 0.294&0.046&0.147 \\
Grounding DINO & 14.107 & 0.060 & 0.105 & 0.190 & 0.102 & 0.048 & 0.878 & \textbf{0.882}&0.107&\textbf{0.324}\\
DyHead & 10.078 & 0.044 & 0.097 & 0.185 & 0.08 & \textbf{0.068}& 0.805 & 0.588 &0.099&0.316\\
YOLOX & 6.016 & 0.023 & 0.059& 0.111& 0.057& 0.014& 0.512 & 0.706 &0.061& 0.204\\
GLIP & \textbf{24.178}& \textbf{0.099}& \textbf{0.106}& \textbf{0.190}& \textbf{0.108}& 0.041 & \textbf{0.927}& \textbf{0.882}&\textbf{0.109}&\textbf{0.324}\\
\midrule

\multicolumn{11}{c}{Rainy night under streetlights} \\
Deformable DETR &4.129 & 0.016& 0.027& 0.046 & 0.032 & 0.0 & 0.208 & 0.583&0.03& 0.095\\
Grounding DINO & 10.167 & 0.033 & 0.044 & 0.071 & 0.054 & 0.0 & 0.333 & \textbf{0.917} &0.047& 0.140 \\
DyHead & 5.542 & 0.023 & 0.048& \textbf{0.082}& 0.050& \textbf{0.012}& 0.417 & 0.667 &0.051& \textbf{0.158}\\
YOLOX & 2.962 & 0.012 & 0.020 & 0.039 & 0.021 & 0.0 & 0.083 & 0.667 &0.024& 0.082\\
GLIP & \textbf{13.588}& \textbf{0.046}& \textbf{0.049}& \textbf{0.082}& \textbf{0.057}& 0.0 & \textbf{0.458}& \textbf{0.917} & \textbf{0.052}& \textbf{0.158}\\
\bottomrule
\end{tabular}
}
\end{table*}

\begin{table*}[htpb]
  \caption{Evaluation results of benchmarks: Instance segmentation, Vehicle.}
  \label{eval results_segmentation_car}
  \centering
  \footnotesize
  \makebox[\linewidth][c]{
  \begin{tabular}{lcccccccccc}
    \toprule
    Model & aPCD (m) & Mean of IoU $\times$ Confidence score & AP$_{50:95}$ &AP$_{50}$ & AP$_{75}$ & AP$_{S}$ &AP$_{M}$ & AP$_{L}$  & AR&F1$_{50}$\\
    \midrule
    \multicolumn{11}{c}{Clear daylight}\\
    Mask R-CNN ok& 89.750& 0.405&   0.376&0.579& 0.410& 0.444&0.942& \textbf{0.947} & 0.381&0.736\\
    ConvNeXt-V2& 89.454& 0.403&   0.395&0.553& 0.453& 0.401&\textbf{0.980}& \textbf{0.947} & 0.399&0.715\\
    SOLOv2& 36.572& 0.165&   0.233&0.276& 0.276& 0.028&\textbf{0.980}& \textbf{0.947} & 0.237&0.438\\
    Mask2Former& \textbf{107.095}& \textbf{0.479}&   \textbf{0.423}&\textbf{0.633}& \textbf{0.492}& \textbf{0.507}&\textbf{0.980}& \textbf{0.947} & \textbf{0.427}&\textbf{0.778}\\
    RTMDet& 43.471& 0.210&   0.349&0.593& 0.338& 0.454&\textbf{0.980}& \textbf{0.947} & 0.353&0.747\\
    \midrule

    \multicolumn{11}{c}{Rainy daylight}\\
    Mask R-CNN
& 67.237& 0.250&   0.249&0.381& 0.246& 0.292&0.563& 0.894 & 0.252&0.556\\
    ConvNeXt-V2
& 78.806& 0.316&   0.352&0.504& 0.410& 0.353&\textbf{0.981}& \textbf{0.947} & 0.355&0.673\\
    SOLOv2
& 24.957& 0.099&   0.144&0.201& 0.157& 0.090&0.399& \textbf{0.947} & 0.147&0.340\\
    Mask2Former
& \textbf{107.037}& \textbf{0.421}&   \textbf{0.380}&\textbf{0.570}& \textbf{0.429}& \textbf{0.452}&0.927& \textbf{0.947} & \textbf{0.383}&\textbf{0.729}\\
    RTMDet& 35.252& 0.152&   0.212&0.365& 0.201& 0.185&0.945& 0.894 & 0.215&0.539\\
    \midrule

    \multicolumn{11}{c}{Rainy night}\\
    Mask R-CNN& 19.096&0.090 &  0.075&0.153& 0.072& 0.0&0.500&0.778  &0.078 &0.269\\
    ConvNeXt-V2& \textbf{24.870}& 0.128&  \textbf{0.094}&\textbf{0.225}& 0.072& \textbf{0.026}&\textbf{0.827}& 0.806 & \textbf{0.096} & \textbf{0.371}\\
    SOLOv2& 11.661& 0.069& 0.090&0.186&0.075 & 0.0&0.673& 0.833 & 0.092&0.318\\
    Mask2Former& 22.170& 0.108& 0.090  &0.186&0.075& 0.011&0.673& 0.750&0.092&0.318\\
    RTMDet& 14.079& 0.080& 0.089  &0.178&\textbf{0.081}& 0.0&0.577&\textbf{0.917} & 0.091&0.306\\

    \midrule
    \multicolumn{11}{c}{Rainy night under streetlights}\\
    Mask R-CNN& 23.587& 0.095& 0.073  &0.152& 0.068&0.032 &0.500& 0.600& 0.076& 0.268\\
    ConvNeXt-V2& \textbf{51.942}&\textbf{0.223} & \textbf{0.160}&\textbf{0.319}& \textbf{0.152}& \textbf{0.151}&\textbf{0.929}& 0.800& \textbf{0.163}&\textbf{0.487}\\
    SOLOv2& 12.976& 0.062&  0.069 &0.121& 0.071& 0.0&0.357&0.767  & 0.072&0.220\\
    Mask2Former& 30.648& 0.136&  0.104 &0.186& 0.115&0.012 &0.786& 0.733 &0.106 &0.318\\
    RTMDet& 21.757& 0.098& 0.110   & 0.207 & 0.121 & 0.020 & 0.833& \textbf{0.833} &0.113 &0.348\\

    \bottomrule
  \end{tabular}
  }
\end{table*}

\begin{table*}[htpb]
  \caption{Evaluation results of benchmarks: Instance segmentation, Pedestrian.}
  \label{eval results_segmentation_person}
  \centering
  \footnotesize
  \makebox[\linewidth][c]{
  \begin{tabular}{lcccccccccc}
    \toprule
    Model & aPCD (m) & Mean of IoU $\times$ Confidence score & AP$_{50:95}$ &AP$_{50}$ & AP$_{75}$ & AP$_{S}$ &AP$_{M}$ & AP$_{L}$  & AR&F1$_{50}$\\
    \midrule
    \multicolumn{11}{c}{Clear daylight}\\
    Mask R-CNN& 21.091& 0.078& 0.048  & 0.133 &0.012 &0.094 &0.565& 0.100& 0.050&0.239\\
    ConvNeXt-V2& 23.448& 0.092& \textbf{0.068}  &\textbf{0.175}& \textbf{0.021}& 0.101& \textbf{0.826} &\textbf{0.700}& \textbf{0.070}&\textbf{0.303}\\
    SOLOv2& 9.261&0.039 &0.040&0.103& 0.015& 0.044&0.565&0.600&0.043 &0.191\\
    Mask2Former& \textbf{23.547}& \textbf{0.096}& 0.055  &0.154&\textbf{0.021} & \textbf{0.104} & 0.652& 0.300& 0.057&0.272\\
    RTMDet& 5.941&0.025 & 0.033  &0.069& 0.030& 0.017& 0.478 & 0.500& 0.035& 0.135\\
    \midrule
    \multicolumn{11}{c}{Rainy daylight}\\
    Mask R-CNN
& 2.238& 0.010&0.006   &0.017&0.003&0.011 &0.0& 0.091&0.008 &0.039\\
    ConvNeXt-V2
&\textbf{24.364} &\textbf{0.108} &  \textbf{0.079} &\textbf{0.212}& \textbf{0.020}& \textbf{0.147}&\textbf{0.538} &\textbf{0.818} & \textbf{0.082}&\textbf{0.354}\\
    SOLOv2
& 3.479&0.020 &0.021   &0.056& 0.010& 0.026&0.115& 0.455& 0.024& 0.112 \\
    Mask2Former
& 15.942& 0.072 & 0.037  &0.096& 0.017& 0.049&0.231& 0.727 & 0.039&0.181\\
    RTMDet&3.162 & 0.018& 0.022  &0.060& 0.003&0.026 &0.269& 0.182& 0.024&0.118\\
    \midrule
    
    \multicolumn{11}{c}{Rainy night}\\
    Mask R-CNN& 0.0&0.001&  0.0&0.0&0.0 &0.0 &0.0&0.0&0.0 &0.0\\
    ConvNeXt-V2& 0.0 & 0.001&  0.0&0.0&0.0 &0.0 &0.0&0.0&0.0 &0.0\\
    SOLOv2& 0.0 & 0.001&  0.0&0.0&0.0 &0.0 &0.0&0.0&0.0 &0.0\\
    Mask2Former& 0.0 & 0.002& 0.0&0.0&0.0 &0.0 &0.0&0.0&0.0 &0.0\\ 
    RTMDet& 0.0 & 0.0&0.0 &0.0 &0.0 &0.0& 0.0 &0.0 &0.0 &0.0\\ 

    \midrule
    \multicolumn{11}{c}{Rainy night under streetlights}\\
    Mask R-CNN& 0.634&0.017& 0.030&0.062& 0.031&0.0&0.0& 0.417 &0.038 &0.136\\
    ConvNeXt-V2& 2.879& 0.028 & 0.012& 0.042 & 0.0&0.0 & 0.083 & 0.167 & 0.018 &0.099\\
    SOLOv2& 1.423& 0.025& \textbf{0.061} &0.094& \textbf{0.073}& 0.0 & \textbf{0.250} & 0.417 & 0.070& 0.189\\
    Mask2Former& \textbf{4.955}&\textbf{0.038} & 0.030 &0.062&0.031 &0.0&\textbf{0.250} &0.167 &0.038 &0.136\\
    RTMDet&3.427 &0.029 &0.059&\textbf{0.104}&\textbf{0.073}&0.0 &0.0& \textbf{0.833} & \textbf{0.068} &\textbf{0.206}\\

    \bottomrule
  \end{tabular}
  }
\end{table*}









\section{Benchmark setup and configuration}


Deformable DETR employs a ResNet-50 backbone and uses a two-stage refinement mechanism to improve region proposal quality and localization. In our configuration, we build on the deformable-detr-refine\_r50 COCO recipe and enable the two-stage pipeline explicitly by setting as\_two\_stage=True.


Grounding DINO utilizes a Swin-B backbone initialized with a pretraining image size
of $384\times 384$, uses 128 embedding dimensions, and follows a $[2,\,2,\,18,\,2]$ depth configuration with a window size of~12 and a
drop-path rate of~0.3. The multi-head self-attention stages use $[4,\,8,\,16,\,32]$ heads, matching the Swin-B architecture. The neck is
adapted to process the high-dimensional feature maps (256/512/1024 channels) produced by the backbone. 


DyHead integrates an ATSS detection head with a Swin Transformer backbone and a hybrid FPN–DyHead neck, following the official COCO training recipe. The model uses a Swin-Large backbone pretrained on ImageNet-22K at a resolution of $384\times384$, producing multi-scale feature maps with 384, 768, and 1536 channels. These features are fed into a five-level Feature Pyramid Network (FPN) with 256-channel outputs and subsequently refined using a six-block DyHead module that applies dynamic, scale-aware attention to strengthen spatial and semantic feature fusion. The ATSS head operates on 256-channel inputs with a $1\times 1$ prediction kernel, a single-ratio anchor generator, $\Delta$XYWH bounding-box coding, and employs Focal Loss for classification, GIoU Loss for regression, and a centerness term for improved localization. Training follows the ATSSAssigner with top-$k$=9, and uses a data pipeline that includes multi-scale random resizing between 
$2000\times 480$ and 
$2000\times 1200$, random horizontal flipping, and dataset repetition to stabilize optimization for the large Transformer backbone. Optimization is performed using AdamW with a learning rate of 
$5\times 10^{-5}$, along with parameter-wise weight-decay handling for positional embeddings, relative-bias tables, and normalization layers. During inference, images are resized to 
$2000\times 1200$, and detections are generated using non-maximum suppression with an IoU threshold of 0.6, keeping up to 100 predictions per image.


YOLOX uses a CSPDarknet backbone (deepen = 0.33, widen = 0.50), a YOLOX-PAFPN neck with 128 output channels, and a decoupled YOLOX head trained on COCO with strong augmentation. In our configuration, the network is scaled up substantially: the backbone deepen and widen factors are increased to 1.33 and 1.25, producing a deeper and wider CSPDarknet. The neck is also enlarged, taking 320/640/1280-channel inputs from the backbone and outputting 320-channel features with four CSP blocks. The YOLOX head is widened accordingly to 320 channels.


GLIP features a Swin-S backbone with 192 embedding dimensions, stage depths of $[2,\,2,\,18,\,2]$, and a window size of~12, together with an increased drop-path rate of~0.4 for stronger regularization. The FPN neck is adapted to process multi-scale features of 384, 768, and 1536 channels from the backbone. 
The ATSS-style bounding box head is enhanced through early fusion of text and visual features and expanded to eight DyHead blocks, with gradient checkpointing enabled to reduce memory consumption. The model is initialized from the publicly available GLIP-L pretrained weights.


Mask~R--CNN is configured with a ResNeXt-101--64$\times$4d backbone and a Feature Pyramid Network (FPN) neck. The backbone uses 101 layers, 64 groups, a base width of~4, four stages with outputs at all stages, and frozen parameters in the first stage. Batch normalization is applied with trainable parameters, and the PyTorch-style configuration is used. The backbone is initialized from the official ResNeXt-101--64$\times$4d pretrained checkpoint. Training follows a multi-scale polynomial learning rate schedule and a 3$\times$ COCO schedule. 


ConvNeXt-V2 is implemented in a Mask~R--CNN framework using a ConvNeXt-V2-B backbone pretrained with FCMAE. The backbone employs the
\texttt{base} configuration with \texttt{out\_indices=[0,1,2,3]}, a drop-path rate of~0.4, GRN enabled, and initialization from the FCMAE checkpoint. An FPN neck processes feature maps of 128, 256, 512, and 1024 channels. Training uses large-scale jittering at a resolution of $1024\times1024$, random cropping, and a 36-epoch schedule with linear warm-up and multi-step learning rate decay. Optimization is performed with AdamW and layer-wise learning rate decay, and testing applies NMS in the RPN and soft-NMS in the RCNN.


SOLOv2 is configured with a ResNeXt-101-64$\times$4d backbone in which DCNv2 deformable convolutions are enabled for stages~2--4, and the
backbone is initialized from the official pretrained ResNeXt-101 checkpoint. An FPN neck processes multi-scale feature maps for instance
segmentation. The mask head employs DCNv2 for all convolutional layers through both the mask feature head and the main mask head configuration.
Training follows the multi-scale strategy defined in the base configuration and uses a 3$\times$ schedule on the COCO dataset.


Mask2Former is configured with a Swin-S backbone using a patch size of~4, a window size of~7, and stage depths of $[2,\,2,\,18,\,2]$, initialized
from the official pretrained weights. The architecture includes a pixel decoder and a transformer decoder head for mask prediction. Training follows a large-scale jittering strategy over 50~epochs with an 8$\times$2 batch configuration. Parameter-wise optimization is applied by assigning reduced learning rates to all backbone layers (\texttt{lr\_mult=0.1}) and zero decay to normalization, positional
embedding, query embedding, and level embedding parameters, as defined in the custom parameter rules.


RTMDet is configured with a CSPNeXt-X backbone using a P5 design, an expand ratio of~0.5, and deepen and widen factors of~1.33 and~1.25, respectively, with integrated channel attention. The CSPNeXtPAFPN neck receives feature maps of 320, 640, and 1280 channels and outputs 320-channel features with four CSP blocks. The RTMDetSepBNHead is configured with 320 input and feature channels and retains the 80-class setting with a DistancePointBBoxCoder. Training uses a base learning rate of~0.002 with a linear warm-up phase for the first 1000~iterations, followed by cosine annealing for the second half of the training
schedule.